\documentclass[letterpaper]{article} 
\usepackage[preprint]{aaai2027}  
\usepackage[hyphens]{url}  
\usepackage{graphicx} 
\usepackage{natbib}  
\usepackage{caption} 
\usepackage{algorithm}
\usepackage{algorithmic}
\usepackage{amsthm}
\usepackage{array}
\usepackage{amssymb}
\usepackage{amsmath}

\usepackage{newfloat}
\usepackage{listings}
\DeclareCaptionStyle{ruled}{labelfont=normalfont,labelsep=colon,strut=off} 
\floatstyle{ruled}
\newfloat{listing}{tb}{lst}{}
\floatname{listing}{Listing}

\lstdefinestyle{instrument}{%
	basicstyle={\footnotesize\ttfamily},
	numbers=none,xleftmargin=0.6em,
	breaklines=true,breakindent=0pt,breakatwhitespace=false,
	columns=fullflexible,keepspaces=true,frame=none,
	aboveskip=3pt,belowskip=3pt,
	showstringspaces=false,tabsize=2,
	extendedchars=true,
	literate={—}{{{\textrm{---}}}}1}
\newcommand{\fs}{\hspace{0.5em}}

\newcommand{\AdjHelpSonnetResult}{%
  \ensuremath{{+}0.500} (\ensuremath{p{=}9.6{\times}10^{-10}})}
\newcommand{\AdjIndepSonnetResult}{%
  \ensuremath{{-}0.302} (\ensuremath{p{=}7.4{\times}10^{-6}})}
\newcommand{\AdjHelpGPTResult}{%
  \ensuremath{{-}0.041} (\ensuremath{p{=}.564})}
\newcommand{\AdjIndepGPTResult}{%
  \ensuremath{{-}0.295} (\ensuremath{p{=}1.7{\times}10^{-6}})}
\newcommand{\AdjHelpGeminiResult}{%
  \ensuremath{{+}0.053} (\ensuremath{p{=}.208})}
\newcommand{\AdjIndepGeminiResult}{%
  \ensuremath{{-}0.604} (\ensuremath{p{=}2.0{\times}10^{-19}})}

\newcommand{\AccMaxSonnet}{\ensuremath{1.00}}
\newcommand{\AccMinSonnet}{\ensuremath{0.77}}
\newcommand{\AgreeHelpKappa}{\ensuremath{.243}}
\newcommand{\AgreeHelpRho}{\ensuremath{.372}}
\newcommand{\AgreeHelpRhoGPT}{\ensuremath{.205}}
\newcommand{\AgreeHelpRhoGemini}{\ensuremath{.585}}
\newcommand{\AgreeHelpRhoSonnet}{\ensuremath{.418}}

\newcommand{\AgreePedKappa}{\ensuremath{.653}}
\newcommand{\AgreePedRho}{\ensuremath{.740}}
\newcommand{\AgreePedRhoGPT}{\ensuremath{.847}}
\newcommand{\AgreePedRhoGemini}{\ensuremath{.431}}
\newcommand{\AgreePedRhoSonnet}{\ensuremath{.754}}
\newcommand{\CeilOpus}{\ensuremath{975}}
\newcommand{\CeilSol}{\ensuremath{1120}}
\newcommand{\DetIndepGPT}{\ensuremath{{-}0.535}}
\newcommand{\DetIndepGPTP}{\ensuremath{2.1{\times}10^{-32}}}
\newcommand{\DetIndepGemini}{\ensuremath{{-}0.592}}
\newcommand{\DetIndepGeminiP}{\ensuremath{9.1{\times}10^{-21}}}
\newcommand{\DetIndepSonnet}{\ensuremath{{-}0.386}}
\newcommand{\DetIndepSonnetP}{\ensuremath{5.6{\times}10^{-12}}}
\newcommand{\DetLeakConvGPT}{\ensuremath{0.766}}
\newcommand{\DetLeakConvGemini}{\ensuremath{0.113}}
\newcommand{\DetLeakConvSonnet}{\ensuremath{0.430}}
\newcommand{\DetLeakGPTD}{\ensuremath{{+}1.00}}
\newcommand{\DetLeakGPTP}{\ensuremath{.0020}}
\newcommand{\DetLeakGeminiD}{\ensuremath{{+}.32}}
\newcommand{\DetLeakGeminiP}{\ensuremath{.625}}
\newcommand{\DetLeakPedGPT}{\ensuremath{0.070}}
\newcommand{\DetLeakPedGemini}{\ensuremath{0.075}}
\newcommand{\DetLeakPedSonnet}{\ensuremath{0.064}}
\newcommand{\DetLeakSonnetD}{\ensuremath{{+}.96}}
\newcommand{\DetLeakSonnetP}{\ensuremath{.0039}}
\newcommand{\HolmHelpGPT}{\ensuremath{9.1{\times}10^{-11}}}
\newcommand{\HolmHelpGemini}{\ensuremath{.354}}
\newcommand{\HolmHelpSonnet}{\ensuremath{3.2{\times}10^{-15}}}
\newcommand{\NratingTotal}{\ensuremath{7074}}
\newcommand{\NturnGPT}{\ensuremath{379}}
\newcommand{\NturnGemini}{\ensuremath{442}}
\newcommand{\NturnSonnet}{\ensuremath{358}}
\newcommand{\NturnTotal}{\ensuremath{1179}}
\newcommand{\OpusHelpGPT}{\ensuremath{{-}0.376}}
\newcommand{\OpusHelpGPTD}{\ensuremath{{-}.74}}
\newcommand{\OpusHelpGPTP}{\ensuremath{.0020}}
\newcommand{\OpusHelpGemini}{\ensuremath{{+}0.088}}
\newcommand{\OpusHelpGeminiD}{\ensuremath{{+}.67}}
\newcommand{\OpusHelpGeminiP}{\ensuremath{.0195}}
\newcommand{\OpusHelpSonnet}{\ensuremath{{-}0.094}}
\newcommand{\OpusHelpSonnetD}{\ensuremath{{-}.10}}
\newcommand{\OpusHelpSonnetP}{\ensuremath{.160}}

\newcommand{\OpusPedConvSonnet}{\ensuremath{2.578}}
\newcommand{\OpusPedGPT}{\ensuremath{{-}2.364}}
\newcommand{\OpusPedGPTD}{\ensuremath{{-}1.00}}
\newcommand{\OpusPedGPTP}{\ensuremath{.0020}}
\newcommand{\OpusPedGemini}{\ensuremath{{-}0.181}}
\newcommand{\OpusPedGeminiD}{\ensuremath{{-}.33}}
\newcommand{\OpusPedGeminiP}{\ensuremath{.426}}

\newcommand{\OpusPedPedSonnet}{\ensuremath{4.299}}
\newcommand{\OpusPedSonnet}{\ensuremath{{-}1.721}}
\newcommand{\OpusPedSonnetD}{\ensuremath{{-}1.00}}
\newcommand{\OpusPedSonnetP}{\ensuremath{.0020}}

\newcommand{\OpusPoolHelpSonnet}{\ensuremath{{+}0.303}}
\newcommand{\OpusPoolHelpSonnetP}{\ensuremath{9.1{\times}10^{-6}}}

\newcommand{\SolAdjHelpGPT}{\ensuremath{{+}0.028}}

\newcommand{\SolAdjHelpGemini}{\ensuremath{{+}0.035}}

\newcommand{\SolAdjHelpSonnet}{\ensuremath{{+}0.044}}

\newcommand{\SolHelpGPT}{\ensuremath{{+}0.086}}
\newcommand{\SolHelpGPTD}{\ensuremath{{+}.70}}
\newcommand{\SolHelpGPTP}{\ensuremath{.0156}}
\newcommand{\SolHelpGemini}{\ensuremath{{+}0.065}}
\newcommand{\SolHelpGeminiD}{\ensuremath{{+}.65}}
\newcommand{\SolHelpGeminiP}{\ensuremath{.0156}}
\newcommand{\SolHelpSonnet}{\ensuremath{{+}0.115}}
\newcommand{\SolHelpSonnetD}{\ensuremath{{+}.84}}
\newcommand{\SolHelpSonnetP}{\ensuremath{.0039}}

\newcommand{\SolPedGPT}{\ensuremath{{-}2.018}}
\newcommand{\SolPedGPTD}{\ensuremath{{-}1.00}}
\newcommand{\SolPedGPTP}{\ensuremath{.0020}}
\newcommand{\SolPedGemini}{\ensuremath{{-}0.016}}
\newcommand{\SolPedGeminiD}{\ensuremath{{+}.05}}
\newcommand{\SolPedGeminiP}{\ensuremath{1.0}}

\newcommand{\SolPedSonnet}{\ensuremath{{-}1.240}}
\newcommand{\SolPedSonnetD}{\ensuremath{{-}.98}}
\newcommand{\SolPedSonnetP}{\ensuremath{.0020}}
\newcommand{\SolPoolHelpGPT}{\ensuremath{{+}0.078}}

\newcommand{\SolPoolHelpGemini}{\ensuremath{{+}0.044}}

\newcommand{\SolPoolHelpSonnet}{\ensuremath{{+}0.096}}

\newcommand{\yes}{$\checkmark$}
\newcommand{\no}{$\times$}
\newcommand{\inc}{$\circ$}
\usepackage{booktabs}

\title{Rethinking LLM-Judged Helpfulness as a Pedagogy Signal: A Pre-Registered Audit Across Tutor Models}
\author{
    Shuyi Fan\textsuperscript{\rm 1}\equalcontrib,
    Boyuan Deng\textsuperscript{\rm 2}\equalcontrib,
    Mengyu Xu\textsuperscript{\rm 3}\equalcontrib,
    Jiale Liu\textsuperscript{\rm 4},
    Hongyang Zhang\textsuperscript{\rm 5},
    Qiaoxin Yang\textsuperscript{\rm 6},
    Chongyang Gao\textsuperscript{\rm 7}\corresponding
}
\affiliations{

    \textsuperscript{\rm 1} Columbia University, USA
    \textsuperscript{\rm 2} Johns Hopkins University, USA
    \textsuperscript{\rm 3} The University of Chicago, USA\\
    \textsuperscript{\rm 4} The Hong Kong University of Science and Technology (Guangzhou)
    \textsuperscript{\rm 5} The Hong Kong Polytechnic University, Hong Kong\\
    \textsuperscript{\rm 6} SynAI Technologies Inc.
    \textsuperscript{\rm 7} Northwestern University, USA\\

    ff2483@tc.columbia.edu, bdeng8@jh.edu, mxu09@uchicago.edu, jialeliu@hkust-gz.edu.cn \\
    hong-yang.zhang@connect.polyu.hk, yangqiaoxinn@qq.com, chongyanggao2026@u.northwestern.edu
}

\begin{document}

\maketitle
\begin{abstract}
     LLM tutoring poses a measurement problem: can a general-purpose helpfulness rubric distinguish direct answer-giving from pedagogical guidance? We audit this signal in a pre-registered study. Within each of three tutor bases, we compare conversational and pedagogical policies instantiated with the same underlying model and paired with one fixed weak simulated student. Deterministic detectors measure answer leakage and next-turn independent work. Claude Opus~4.8 is the frozen, condition-blind primary judge. After the Opus scores were fixed, GPT-5.6~Sol was prospectively specified for a post hoc robustness audit of the same 1{,}179 confirmatory answer-phase tutor turns under the frozen helpfulness and pedagogy rubrics. On the primary base under Opus, the policies do not differ significantly in helpfulness but are perfectly rank-separated under the pedagogy rubric (Cliff's $|\delta|{=}0.10$ vs.\ $1.0$). Across the two judges, pedagogy contrasts retain their direction where detected, whereas the helpfulness ordering is judge-contingent, reversing between judges on two of three bases. In an Opus-only ablation, seven primary-base policies span $2.3$ points in mean judged pedagogy within a $0.25$-point band of mean judged helpfulness. Separately, answer-revealing turns are followed by less independent student work on every base, a result that is judge-invariant by construction. In this controlled setting, general-purpose helpfulness is not a reliable pedagogy signal. Tutor evaluation should pair pedagogy-targeted rubrics with deterministic process measures.
\end{abstract}

\begin{links}
    \link{Code}{https://github.com/bydeng01/conv-vs-ped-tutor}
\end{links}

\section{Introduction}
 
Recent work uses expert preference evaluations and preference-based optimization to develop LLM tutors with stronger pedagogical behavior \citep{learnlmteam2025learnlmimprovinggeminilearning, sonkar2024pedagogical}. One failure mode is well documented: a model tuned to be helpful \citep{ouyang2022training} tends to give a stuck student the answer rather than withhold it \citep{puech2025towards}. In a field experiment, students given an unguarded GPT-4 tutor performed worse on a subsequent unassisted exam than students in the no-tool control \citep{bastani2025generative}. Whether this behavior is caught depends on the evaluation signal, and strong LLMs increasingly serve as scalable proxies for human preference judgments in open-ended assistant evaluation \citep{zheng2023judgingllmasajudgemtbenchchatbot}. Such an evaluator could entrench the behavior by rewarding it, or it could simply fail to see it. We treat this as an empirical question and audit the evaluator directly.
 
Our testbed fixes the base model and varies only the tutoring policy. ConvTutor is a minimal single-call ``helpful tutor'' whose answer-giving is emergent, not engineered. PedTutor routes the same frozen model through a learner-state tracker and then one of three principle-grounded responders. A no-tutor cold baseline completes the design. All three conditions face one deliberately weak simulated student (Llama-3.1-8B~\citep{grattafiori2024llama3herdmodels}, about $11\%$ unaided accuracy), making tutoring load-bearing. We score every answer-phase tutor turn on two evaluation layers. The primary layer uses Claude Opus~4.8 \citep{anthropic2026opus48} as a frozen, condition-blind judge. Opus rates annotator-perceived helpfulness and, post hoc, a symmetric pedagogy rubric. The judge identity, helpfulness rubric, metric definitions, and outcome-interpretation table were fixed before confirmatory data collection. The pedagogy rubric was pre-registered before pedagogy scoring. After freezing the Opus scores, we prospectively specified a post hoc robustness layer. GPT-5.6~Sol~\citep{openai2026gpt56} rescored the identical $1{,}179$ confirmatory tutor turns under the same two rubrics. Two deterministic detectors, answer leakage and next-turn independent work, measure process directly. We ran three tutor bases (Claude Sonnet~4.6, GPT-5.5, and Gemini~3.1~Pro Preview)~\citep{anthropic2026sonnet46,openai2026gpt55,googledeepmind2026gemini31pro}, with ten replicates per condition per base ($90$ sessions).
 
The study yielded four results. First, on the primary base under Opus, the helpfulness contrast was small and nonsignificant (Cliff's $|\delta|{=}0.10$), whereas the pedagogy contrast produced perfect rank separation ($|\delta|{=}1.0$). The pre-registered joint claim that helpfulness would reward answer-giving was therefore not supported. Second, the ConvTutor--PedTutor helpfulness ordering was judge-contingent. It reversed between Opus and Sol on the Sonnet and GPT-5.5 bases and replicated only on Gemini. By contrast, the pedagogy-targeted rubric retained its direction under both judges wherever a policy difference was detected. Third, an Opus-only ablation showed the same helpfulness--pedagogy compression across seven primary-base policies. A prompt-only Socratic instruction therefore reached or exceeded the routed policy's judged-pedagogy score. Fourth, answer-revealing turns were followed by less independent student work on every tutor base. This deterministic coupling is judge-invariant by construction. Together, these findings locate the weakness in the general-purpose helpfulness signal rather than in any single null result.
 
We make four contributions. First, the controlled testbed isolates tutoring policy while holding the base model fixed. Second, it shows that general-purpose helpfulness can miss pedagogical differences and yield a judge-contingent policy ordering (\S\ref{sec:primary}, \S\ref{sec:crossmodel}). Third, deterministic process measures anchor the comparison where the judged signal does not (\S\ref{sec:crossmodel}). Fourth, the findings motivate pairing pedagogy-targeted rubrics with process measures rather than relying on one general-purpose helpfulness score (\S\ref{sec:discussion}). We build no new tutor and make no claim about human learning.

\section{Related Work}
\label{sec:related}

\textbf{Reliability of LLM judges and reward models.} A growing literature documents position, verbosity, self-preference, and other systematic biases in LLM-as-a-judge evaluation, with magnitudes that vary across judges and tasks \citep{zheng2023judgingllmasajudgemtbenchchatbot,wataoka2024self,koo2024benchmarking,shi2025judging}. Meta-evaluation further shows that judge--prompt combinations can fail on adversarial instruction-following pairs whose superficial qualities obscure correctness \citep{zeng2024evaluating}. Reward-model benchmarks likewise expose refusal, reasoning, and instruction-following limitations across chat, safety, and out-of-distribution comparisons \citep{lambert2025rewardbench}. We study a distinct construct-validity concern: whether general-purpose helpfulness captures pedagogically relevant differences between tutors. Comparing two LLM judges bounds robustness across evaluators but does not establish human validity.

\textbf{LLM tutors and pedagogical alignment.} A fast-growing line builds pedagogically tuned tutors: LearnLM's pedagogical instruction-following \citep{learnlmteam2025learnlmimprovinggeminilearning, jurenka2025responsibledevelopmentgenerativeai}, preference-based pedagogical alignment \citep{sonkar2024pedagogical}, and Socratic teaching \citep{liu2024socraticlm}. Structured tutors that constrain answer-giving \citep{pal2024autotutor} are PedTutor's closest antecedents. We build no better tutor; we hold the tutor fixed and audit the evaluator.

\textbf{Intelligent tutoring and the learning sciences.} Scaffolding \citep{Wood1976TheRO}, the generation effect \citep{Slamecka1978TheGE}, the assistance dilemma \citep{koedinger2007exploring}, and tutored help-seeking \citep{Aleven2006TowardMT} all turn on preserving productive struggle. We study this tension with two turn-level measures: whether the tutor discloses the answer and whether the student attempts independent reasoning on the following turn.

\textbf{Simulated students as instruments.} Capable models can fail to emulate weak learners (the competence paradox) and can produce unrealistic error patterns \citep{yuan2026towards, scarlatos2026simulated}. We therefore use a genuinely weak open model as a controlled instrument rather than a strong model role-playing weakness.

\section{Study Design and Pre-Registered Inference}
\label{sec:design}
 
\subsection{Testbed: Two Policies, Identical Weights}
\label{sec:design-testbed}
 
\begin{figure*}[t]
    \centering
    \includegraphics[width=0.85\textwidth]{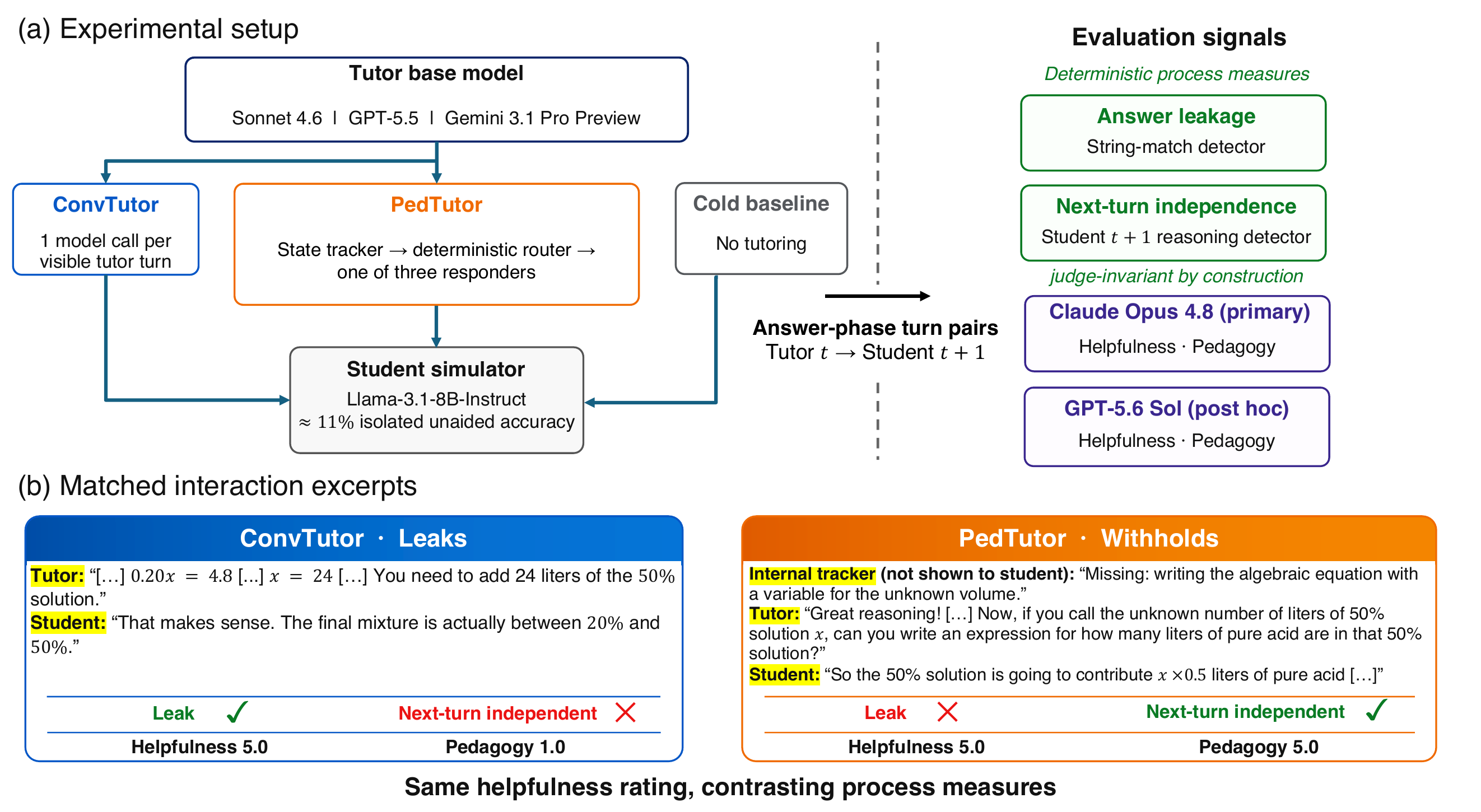}
    \caption{Study design and matched interaction excerpts. (a) Within each tutor base, ConvTutor and PedTutor use identical base weights. ConvTutor uses one model call per visible turn; PedTutor uses a learner-state tracker and one of three routed responders, requiring two calls. A no-tutoring cold baseline and weak student simulator complete the design. Answer leakage and next-turn independence are deterministic. Claude Opus~4.8 is the condition-blind primary judge for both rubrics; GPT-5.6 Sol is a post hoc robustness judge over the same frozen transcripts. (b) An illustrative matched pair from \texttt{train-1} (replicate 0, first answer-phase tutor turn). Opus gives both turns a helpfulness score of $5/5$, whereas their process measures and pedagogy ratings differ ($1.0$ vs.\ $5.0$). Brackets mark omissions; mathematical notation is normalized for readability.}
    \label{fig:design}
\end{figure*}

Within each tutor base, ConvTutor and PedTutor are instantiated from \emph{identical base weights}; the manipulation is confined to the policy layer (control flow and prompts) and never touches the weights (Figure~\ref{fig:design}a). \textbf{ConvTutor} is a minimal conversational tutor: one realistic ``helpful tutor'' instruction, one model call per turn. Its answer-giving is the phenomenon under study, not engineered. \textbf{PedTutor} is a four-node routed policy over the same base model: a learner-state tracker runs on every visible turn, and a deterministic router then dispatches to exactly one of three responders: a problem decomposer, an answer-deferral gate, or a hint cascade. As a result, each visible tutor turn makes two model calls rather than four sequential stages. The nodes are grounded in contingent tutoring, the assistance dilemma, the generation effect, and tutored help-seeking (citations in Related Work). A \textbf{cold} baseline completes the design: the identical probe schedule with no tutoring in any phase.

The student is deliberately weak so that tutoring is load-bearing: Llama-3.1-8B-Instruct, with no memory beyond its growing context window and no access to reference solutions. Evaluated on the problems in isolation, it solved roughly 11\% of the mixture and weighted-average problems, and its transcripts showed failures to formulate correct setups. Each session is one continuous protocol: six tutored training problems, then three immediate, four interference, three delayed, and three transfer problems, all unassisted probes under the session's accumulated context. The primary base is Claude Sonnet~4.6. We pre-registered two replications that changed the tutor base model to GPT-5.5 or Gemini~3.1~Pro Preview; the student, judge, problems, prompts, evaluation window, and measures remained fixed. The serving vendor co-varies, and Gemini's provider enforces a minimum reasoning setting, so no base is a pure base-model swap. Ten replicates per condition per base give 90 confirmatory sessions. The replicate fixes problem order, condition order, and the student's initial context, and it is the pre-registered unit of pairing; residual sampling stochasticity is within-replicate noise.
 
\subsection{Frozen Process Measures}
\label{sec:design-measures}
 
All process measures are computed over each training problem's answer phase: the exchange up to the student's first committed final answer. Post-resolution conversation counts toward neither tutor, under a rule fixed before confirmatory collection and applied identically to both conditions. Cost and probe accuracy use the full session.

Two prespecified deterministic measures capture answer disclosure and subsequent student reasoning. \textbf{Answer leakage} is the fraction of answer-phase tutor turns that reveal the current problem's solution, by string-level matching against numeric and solution forms declared per problem in advance. Paraphrased reveals are deliberately not counted, making leakage a conservative lower bound. \textbf{Student independence} scores, for each answer-phase student turn, whether it attempts a reasoning step (an algebraic expression or numerical computation) before the tutor speaks again. It enters the analysis at two distinct levels: the session-level \emph{independence ratio} (P3), the fraction of a session's answer-phase student turns that attempt reasoning, and the turn-level \emph{next-turn independence} (J2), the binary status of the first student turn after a given tutor turn. Both labels are generated solely by these deterministic rules; no LLM-based relabeling was used, so both are judge-invariant by construction.
 
\textbf{Annotator-perceived helpfulness} tests whether a standard helpfulness evaluation captures pedagogically relevant behavior. The condition-blind primary judge (Claude Opus~4.8) rates each answer-phase tutor turn three times on a five-point rubric for clarity, responsiveness, and helpfulness, without seeing reference solutions or condition labels; we summarize both the mean and variance across ratings. The score is a reward-model-style proxy for annotator preference; it measures neither the learner's experience nor pedagogical quality.
\textbf{Judged pedagogy} is a second, pre-registered rubric applied post hoc by the same judge to the same turns under the same repetition scheme: four subscores on the four principles above plus a holistic five-point rating. The rubric is written symmetrically, crediting a scaffolding ConvTutor turn as readily as it penalizes a vague PedTutor one. Because PedTutor is built on the same principles this rubric scores, any pedagogy gap between the tutors is reported as a manipulation check rather than independent evidence. The construction-independent signal is the divergence between judged helpfulness and judged pedagogy on the same turns (\S\ref{sec:primary}).
Accuracy is secondary because the simulated student accumulates worked examples in context, causing delayed and transfer performance to approach ceiling and preventing its interpretation as durable learning.
 
\subsection{Pre-Registered Inference}
\label{sec:design-inference}
 
The confirmatory analysis, with the pre-committed decision rules of Table~\ref{tab:preregistered}, was frozen before any confirmatory data existed. Session-level marginals P1 (leakage, ConvTutor${>}$PedTutor), P2 (judged helpfulness, ConvTutor${>}$PedTutor), and P3 (independence, PedTutor${>}$ConvTutor) are tested with two-sided Wilcoxon signed-rank tests paired by replicate at $\alpha{=}.05$, with Cliff's $\delta$ and bootstrap 95\% CIs; pooled turn-level averages are never used for inference. P1 and P3 are manipulation checks, being partly induced by the policy design; P2 is the live test. \textbf{J1} is supported only when P1, P2, and P3 are each significant in the pre-registered direction. \textbf{J2}, the pre-registered headline, is a per-turn coupling fit within each base over the pooled ConvTutor and PedTutor answer-phase turns fixed by the plan ($n{=}\NturnSonnet{}$ on the primary base):
\[
\begin{aligned}
Y_{krpt} &= \beta_0 + \beta_L L_{krpt} + u_r + v_p + \varepsilon_{krpt},\\
Y_{krpt} &= \beta_0 + \beta_L L_{krpt} + \beta_C C_k + u_r + v_p + \varepsilon_{krpt}.
\end{aligned}
\]
Here $k\in\{\mathrm{ConvTutor},\mathrm{PedTutor}\}$ indexes tutoring policy, $L_{krpt}$ is binary answer leakage on turn $t$ of problem $p$ in replicate $r$, and $C_k{=}1$ for PedTutor and $0$ for ConvTutor. Terms $u_r$ and $v_p$ are crossed replicate and problem random intercepts. The outcome $Y_{krpt}$ is either the three-rating mean helpfulness score or the next-turn independence status of the first subsequent student turn. The first line is the pre-registered pooled specification. The second adds tutoring policy $C_k$ as a fixed effect, providing an additive post hoc sensitivity analysis without replacing the pre-registered model. We fit both specifications separately by tutor base and, for helpfulness, by judge. For binary next-turn independence, $\beta_L$ is a linear-probability risk difference rather than an odds ratio. J2 is supported only when $\beta_L$ is significantly positive for helpfulness and significantly negative for next-turn independence.

\textbf{Matched visible-turn budget.} The policies need not spend equal visible turns. PedTutor used $22.3$ in-window tutor turns per session against ConvTutor's $13.5$ (paired gap ${+}8.8$, 95\% CI $[{+}4.6, {+}12.6]$, eight of ten replicate pairs). A matched visible-turn re-analysis, its rule fixed after the full-window result was read, equalizes them. For each problem--replicate pair we set $K{=}\min(n_{\mathrm{Conv}},n_{\mathrm{Ped}})$ over in-window tutor turns, keep the first $K$ in both conditions with the student turns through the response to turn $K$, and recompute the marginals and both J2 legs under the same aggregation and inference. The rule is symmetric across conditions and drops the five problem--replicate cells where $K{=}0$.

\begin{table}[t]
\centering
\small
\setlength{\tabcolsep}{3pt}
\begin{tabular}{@{}l >{\raggedright\arraybackslash}p{4.50cm} ccc@{}}
\toprule
& & \multicolumn{3}{c}{\textbf{Verdict by base}}\\
\cmidrule(l){3-5}
\textbf{Claim} & \textbf{Pre-registered decision rule} & Son & GPT & Gem\\
\midrule
P1 & Session leakage, Conv${>}$Ped, significant; manipulation check & \yes & \yes & \inc\\
P2 & Session judged helpfulness, Conv${>}$Ped, significant; live test & \no & \no & \yes\\
P3 & Session independence ratio, Ped${>}$Conv, significant; manipulation check & \inc & \yes & \inc\\
J1 & P1, P2 and P3 each significant in its pre-registered direction & \no & \no & \no\\
J2 & Leakage${\to}$helpfulness ${>}0$ \emph{and} leakage${\to}$next-turn independence ${<}0$, both significant & \yes & \no & \no\\
\bottomrule
\end{tabular}
\caption{Pre-registered claims, their frozen decision rules, and verdicts by base (\yes\ supported; \no\ not supported; \inc\ inconclusive, a manipulation check that did not separate). Among the component tests, only P2 and J2's helpfulness leg use the frozen Opus judge; P1, P3, and J2's independence leg are deterministic. The J1 and J2 verdicts inherit judge dependence through their helpfulness components. P1--P3 and the J2 legs carry no multiplicity correction. The pedagogy rubric, ablation, and Sol audit do not enter these verdicts.}
\label{tab:preregistered}
\end{table}

\textbf{Cross-judge robustness audit.} After freezing the Opus scores, but before conducting the second-judge analysis, we specified GPT-5.6~Sol for a post hoc robustness audit. Sol rescored the identical $1{,}179$ confirmatory ConvTutor and PedTutor answer-phase tutor turns (\NturnSonnet{} Sonnet, \NturnGPT{} GPT-5.5, \NturnGemini{} Gemini). Each turn received three ratings per rubric, yielding \NratingTotal{} ratings across the frozen helpfulness and pedagogy rubrics. We held the dialogue reconstruction, prompts, rubrics, parser, aggregation rule, answer-phase window, and condition blinding fixed. The judge changed, along with four serving-side aspects (Appendix~\ref{sec:est-crossjudge}). We aligned the two judges' per-turn aggregates by exact turn identity. For instrument $m\in\{\text{helpfulness},\text{pedagogy}\}$, we defined the cross-judge score difference as $D_{krpt}^{(m)}{=}S^{(m)}_{krpt,\mathrm{Sol}}-S^{(m)}_{krpt,\mathrm{Opus}}$, where $S^{(m)}$ is the three-rating mean. Fitting the policy-adjusted structure above to $D_{krpt}^{(m)}$ produced the judge-by-leakage and judge-by-policy interaction terms. This difference is defined only for judged scores. Sol therefore neither rescored nor replicated the deterministic leakage and next-turn independence outcomes. We applied Holm correction within three declared families of three base-wise tests. We reported the judges separately and never averaged them into a consensus.
 
\textbf{Epistemic status.}
Leakage is not randomized at the turn level, so J2 is an observational coupling within pre-assigned policies, not a causal estimate, and part of any next-turn drop is proximal answer adoption (the leaked answer sitting in the student's context). We therefore rest its interpretation on non-definitional properties: the independence leg varies more than threefold across tutor policies (\S\ref{sec:ablation}) and survives the matched visible-turn-budget re-analysis (${+}0.295$ and ${-}0.355$, both significant). With ten paired replicates, session-level nulls rule out only large effects ($d\gtrsim1$), hence the effect-size-contrast form of the evidence in \S\ref{sec:primary}.

\section{Primary-Base Results}
\label{sec:primary}
 
\noindent\textbf{Manipulation checks and the joint claim.} The leakage manipulation fires: ConvTutor's mean per-session leakage fraction is 43.0\% against 6.4\% for PedTutor (P1 supported: paired difference ${+}0.365$, 95\% CI $[{+}0.217, {+}0.516]$, nine of ten replicate pairs; Table~\ref{tab:crossmodel}). Independence moves in the predicted direction but does not separate (P3 inconclusive: difference ${-}0.076$, CI $[{-}0.189, {+}0.021]$). Under Opus, the helpfulness judge did not prefer ConvTutor (P2 not supported: difference \OpusHelpSonnet{}, CI $[{-}0.251, {+}0.058]$, $p{=}\OpusHelpSonnetP{}$). J1 was therefore not supported: at the session level, answer-giving was not rewarded. Probe accuracy saturated (\AccMinSonnet{} to \AccMaxSonnet{} across immediate, delayed, and transfer problems in all three conditions), so it cannot distinguish them.
 
\begin{figure}[t]
    \centering
    \includegraphics[width=0.80\columnwidth]{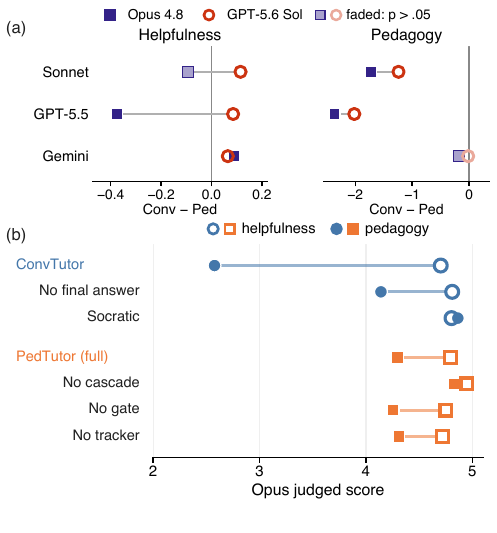}
    \caption{(a) Means of ten paired session-level ConvTutor$-$PedTutor differences in judged helpfulness (left) and judged pedagogy (right) for each tutor base under Claude Opus~4.8 (filled squares) and GPT-5.6~Sol (open circles); faded markers denote $p>.05$ in the two-sided paired Wilcoxon test. In the helpfulness panel, the mean ordering reverses between judges on Sonnet and the GPT-5.5 base, although the Opus Sonnet gap is not detected; both judges favor ConvTutor on Gemini. The pedagogy contrasts retain their sign across judges wherever detected and are reported as a manipulation check. (b) Replicate-averaged Opus judged-helpfulness (open) and judged-pedagogy (filled) scores for two baselines and five ablation variants on the primary Sonnet base; circles and squares denote the ConvTutor and PedTutor families, respectively. Policy means span $0.25$ points for helpfulness and $2.3$ points for pedagogy; Sol did not rescore these policies.}
    \label{fig:dissociation}
\end{figure}

\medskip\noindent\textbf{Helpfulness and pedagogy diverge under Opus.} Judged pedagogy averages \OpusPedConvSonnet{} for ConvTutor and \OpusPedPedSonnet{} for PedTutor (Figure~\ref{fig:dissociation}a). Per \S\ref{sec:design-measures}, we report this gap as a manipulation check. The separation is perfect at the replicate level ($|\delta|{=}1.0$: every PedTutor replicate scores above every ConvTutor replicate), and the rubric is not floored on ConvTutor.
 
The finding that survives this caveat is the divergence: under Opus, rating the same turns under two rubrics, a helpfulness contrast that does not differ significantly ($|\delta|{=}0.10$) sits beside a maximal pedagogy contrast ($|\delta|{=}1.0$). This is the failure mode that makes a general-purpose helpfulness signal unreliable for selecting between tutors. The dissociation rests on this effect-size contrast, not on a single null.
 
\medskip\noindent\textbf{The per-turn coupling.} Both legs of the pre-registered J2 coupling hold on the primary base ($n{=}\NturnSonnet{}$ answer-phase tutor turns). Under Opus, a turn that leaks the answer predicts a higher helpfulness rating (coefficient \OpusPoolHelpSonnet{}, 95\% CI $[{+}0.169, {+}0.437]$, $p{=}\OpusPoolHelpSonnetP{}$); deterministically, it is followed by less independent student work on the following turn (\DetIndepSonnet{}, CI $[{-}0.495, {-}0.276]$, $p{=}\DetIndepSonnetP{}$). Descriptively, leaky turns are rated $4.95$ versus $4.67$ and are followed by a reasoning attempt $33\%$ versus $73\%$ of the time; a replicate-clustered sensitivity gave the same directions. Behaviors invisible to the session-level marginals therefore couple, turn by turn, to Opus ratings and to subsequent student work.

The coupling survives the matched visible-turn-budget re-analysis of \S\ref{sec:design-inference}: both associations persisted ($n{=}256$; ${+}0.295$, $p{=}2.4{\times}10^{-4}$; ${-}0.355$, $p{=}7.5{\times}10^{-8}$). The session marginals kept their direction and their verdicts under the same re-analysis. P1 was unchanged ($p{=}.0039$), while both nonsignificant contrasts weakened further (P2 $p{=}.160$ to $.426$; P3 $p{=}.375$ to $.695$). The matched window also left PedTutor with six leaky turns, below the ten the sensitivity analysis adopts for an adequate per-condition cell. Under that window the J2 legs therefore rested mainly on ConvTutor turns. Whether the helpfulness ordering survives a second evaluator is the question we take up next.
 
\section{Cross-Base Replication and Judge Audit}
\label{sec:crossmodel}
 
\noindent\textbf{The deterministic anchor.} The two replications varied the tutor's base model, with the serving vendor and Gemini's reasoning setting co-varying (\S\ref{sec:design-testbed}; Table~\ref{tab:crossmodel}). Leakage predicted less independent student work on every base ($\beta$ from \DetIndepSonnet{} to \DetIndepGemini{}, all $p{<}10^{-11}$; Table~\ref{tab:crossmodel}b). A replicate-clustered sensitivity analysis on this leg yielded comparable estimates on all three bases (all $p{\leq}.006$). This coupling is deterministic (\S\ref{sec:design-measures}), so the Sol audit could not rescore it.
 
\medskip\noindent\textbf{Frozen Opus cross-base results.} Under the frozen Opus judge, the session-level helpfulness ordering differed by base (Table~\ref{tab:crossmodel}a). P2 was null on Sonnet, significantly favored PedTutor on the GPT-5.5 base, and significantly favored ConvTutor on Gemini. J1 was not supported on any base. After adjusting for tutoring policy, the coefficient was positive on Sonnet (\AdjHelpSonnetResult{}) but not detectably different from zero on GPT-5.5 (\AdjHelpGPTResult{}) or Gemini (\AdjHelpGeminiResult{}). The helpfulness leg was also method-sensitive. Under the replicate-clustered sensitivity the pooled leakage coefficient was ${-}0.267$ on GPT-5.5 ($p{=}.105$), so the significant Opus reversal there did not survive clustering. On Gemini it was ${+}0.066$ ($p{=}.0020$), an association the pre-registered mixed model did not detect, so under that method J2 would be supported there. Table~\ref{tab:preregistered} reports the pre-registered mixed-effects verdicts; we treat the clustered fit as a sensitivity analysis. The policy-adjusted leakage-to-independence coefficient remained negative throughout (\AdjIndepSonnetResult{}, \AdjIndepGPTResult{}, \AdjIndepGeminiResult{}). Under the pedagogy rubric, Opus favored PedTutor wherever it detected a difference and detected none on Gemini (Table~\ref{tab:crossmodel}a).
 
\begin{table*}[!t]
\centering
{\small
(a) Judged policy gaps by tutor base and judge (ConvTutor $-$ PedTutor; both rubrics scored 1--5; signed-rank paired by replicate, 10 pairs, with zero differences omitted from the statistic; $\delta$ is unpaired between-condition dominance)\\[3pt]
\setlength{\tabcolsep}{1.4mm}
\begin{tabular}{@{}ll rrl rrl@{}}
\toprule
 & & \multicolumn{3}{c}{\textbf{Helpfulness} gap} & \multicolumn{3}{c}{\textbf{Pedagogy} gap$^{\ddagger}$} \\
\cmidrule(lr){3-5}\cmidrule(l){6-8}
Tutor base & Judge & $\Delta$ & $\delta$ & $p$ & $\Delta$ & $\delta$ & $p$ \\
\midrule
Sonnet 4.6 & Opus & \OpusHelpSonnet & \OpusHelpSonnetD & \OpusHelpSonnetP & \OpusPedSonnet & \OpusPedSonnetD & \OpusPedSonnetP \\
(primary) & Sol & \SolHelpSonnet & \SolHelpSonnetD & \SolHelpSonnetP & \SolPedSonnet & \SolPedSonnetD & \SolPedSonnetP \\
\addlinespace[2pt]
GPT-5.5 & Opus & \OpusHelpGPT & \OpusHelpGPTD & \OpusHelpGPTP & \OpusPedGPT & \OpusPedGPTD & \OpusPedGPTP \\
 & Sol & \SolHelpGPT & \SolHelpGPTD & \SolHelpGPTP & \SolPedGPT & \SolPedGPTD & \SolPedGPTP \\
\addlinespace[2pt]
Gemini 3.1 & Opus & \OpusHelpGemini & \OpusHelpGeminiD & \OpusHelpGeminiP & \OpusPedGemini & \OpusPedGeminiD & \OpusPedGeminiP \\
Pro prev.$^{\S}$ & Sol & \SolHelpGemini & \SolHelpGeminiD & \SolHelpGeminiP & \SolPedGemini & \SolPedGeminiD & \SolPedGeminiP \\
\bottomrule
\end{tabular}

\vspace{3pt}
(b) Deterministic process by tutor base (judge-invariant by construction)\\[3pt]
\setlength{\tabcolsep}{1.4mm}
\begin{tabular}{@{}l cccc ccc@{}}
\toprule
 & \multicolumn{4}{c}{\textbf{Answer leakage} (frac.)} & \multicolumn{3}{c}{\textbf{Leakage} ${\to}$ \textbf{next-turn indep.}} \\
\cmidrule(lr){2-5}\cmidrule(l){6-8}
Tutor base & conv & ped & $\delta$ & $p$ & $n$ & $\beta$ & $p$ \\
\midrule
Sonnet 4.6 & \DetLeakConvSonnet & \DetLeakPedSonnet & \DetLeakSonnetD & \DetLeakSonnetP & \NturnSonnet & \DetIndepSonnet & \DetIndepSonnetP \\
GPT-5.5 & \DetLeakConvGPT & \DetLeakPedGPT & \DetLeakGPTD & \DetLeakGPTP & \NturnGPT & \DetIndepGPT & \DetIndepGPTP \\
Gemini 3.1 Pro prev.$^{\S}$ & \DetLeakConvGemini & \DetLeakPedGemini & \DetLeakGeminiD & \DetLeakGeminiP & \NturnGemini & \DetIndepGemini & \DetIndepGeminiP \\
\bottomrule
\end{tabular}
}
\caption{Cross-judge policy contrasts; tutor bases are analyzed separately, and the two judges are reported separately, never averaged. (a) Judged policy gaps $\Delta$ (ConvTutor minus PedTutor) for helpfulness and pedagogy. Because $\delta$ ignores the pairing, it can differ in sign from $\Delta$ (Gemini/Sol pedagogy). Helpfulness contrasts reverse in sign between judges on Sonnet and GPT-5.5 (\S\ref{sec:crossmodel}). The pedagogy gap retains its sign wherever a difference is detected and is a manipulation check ($^{\ddagger}$; \S\ref{sec:design-measures}), since PedTutor is built on the principles the rubric scores. (b) Answer-leakage entries are means of the ten session-level fractions per condition, with unpaired Cliff's $\delta$ and a two-sided paired signed-rank $p$. The pre-registered pooled leakage-to-next-turn-independence coefficient $\beta$ is a linear-probability risk difference fitted across ConvTutor and PedTutor turns within each base, using $n$ answer-phase tutor turns. $^{\S}$Gemini uses a provider-enforced minimum reasoning setting absent from the other bases, so cross-base differences cannot be attributed to model family alone.}
\label{tab:crossmodel}
\end{table*}

\begin{figure*}[tb]
    \centering
    \includegraphics[width=0.85\textwidth]{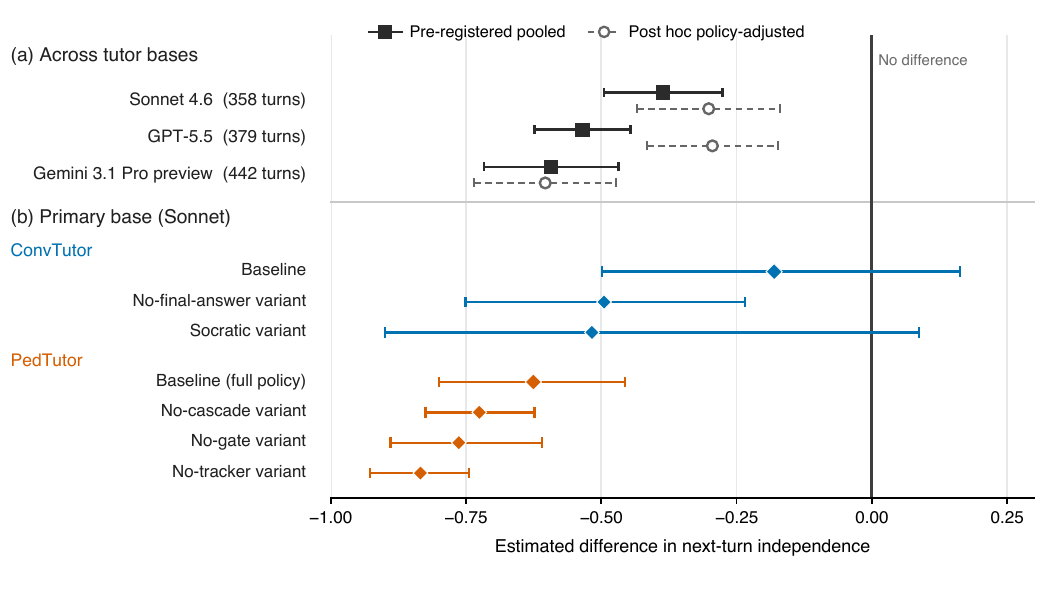}
    \caption{Answer leakage versus next-turn student independence (answer-phase window): a deterministic outcome that Sol never rescored. (a) Pre-registered pooled (filled) and post hoc policy-adjusted (open) leakage coefficients across tutor bases, with model-based 95\% CIs. (b) Descriptive within-replicate leaky-minus-nonleaky differences for the seven Sonnet policies (percentile-bootstrap 95\% CIs). Every estimate is negative; two policy intervals include zero.}
    \label{fig:forest}
\end{figure*}
 
\medskip\noindent\textbf{Sol cross-base results.} Sol's session-level helpfulness contrasts were positive on all three bases (ConvTutor${>}$PedTutor; Table~\ref{tab:crossmodel}a). Under the pedagogy rubric, Sol retained the Opus direction wherever a difference was detected and, like Opus, detected none on Gemini. Under the pre-registered pooled specification its leakage-to-helpfulness coefficient was positive on all three bases (\SolPoolHelpSonnet{}, \SolPoolHelpGPT{}, \SolPoolHelpGemini{}; the first two significant at $p{<}.01$), so on GPT-5.5 the two judges' pooled coefficients are significant with opposite signs. Adjusted for policy, it was not detectably different from zero anywhere (\SolAdjHelpSonnet{}, \SolAdjHelpGPT{}, \SolAdjHelpGemini{}; all raw $p{>}.2$).
 
\medskip\noindent\textbf{The helpfulness ordering was judge-contingent.} The ConvTutor--PedTutor helpfulness ordering reversed between Opus and Sol on two of three bases (Table~\ref{tab:crossmodel}a). On Sonnet, the Opus contrast was negative and nonsignificant, whereas the Sol contrast was positive and significant. This was an ordering reversal, not a significant reversal. On the GPT-5.5 base, the two contrasts were significant and opposite (Opus \OpusHelpGPT{} vs.\ Sol \SolHelpGPT{}). Both deterministic manipulation checks separated the policies on that base under either judge, so the pre-registered joint claim there turned on its single judged component. J1 was therefore not supported under Opus but supported under Sol. No other base flipped: P3 did not separate on Sonnet, and neither P1 nor P3 separated on Gemini, under both judges. Table~\ref{tab:preregistered} reports the frozen Opus verdicts throughout. On Gemini, both judges favored ConvTutor, providing a narrow replication. The judge-by-policy interaction on helpfulness was significant on Sonnet and GPT-5.5 but not on Gemini (Holm-adjusted $p{=}\HolmHelpSonnet{}$, \HolmHelpGPT{}, and \HolmHelpGemini{}). The nonsignificant Gemini interaction does not establish judge equivalence. Independently, every deterministic leakage--independence estimate was negative across three bases and all seven ablation policies in \S\ref{sec:ablation} (Figure~\ref{fig:forest}).
 
\medskip\noindent\textbf{Scale, ceiling, and same-family caveats.} Four caveats bound this result. Cross-judge agreement varied by base and inverted on Gemini. On Sonnet and GPT-5.5 the judges agreed more closely on pedagogy than on helpfulness (Spearman $\rho{=}\AgreePedRhoSonnet{}$ and \AgreePedRhoGPT{}, against \AgreeHelpRhoSonnet{} and \AgreeHelpRhoGPT{}). On Gemini that order reversed ($\rho{=}\AgreePedRhoGemini{}$ against \AgreeHelpRhoGemini{}). Helpfulness agreement was weakest on GPT-5.5, the one base where the two judges' contrasts were significant and opposite. Pooled across bases, agreement reads high on pedagogy ($\rho{=}\AgreePedRho{}$, quadratic-weighted $\kappa{=}\AgreePedKappa{}$) and modest on helpfulness ($\rho{=}\AgreeHelpRho{}$, $\kappa{=}\AgreeHelpKappa{}$). That summary masks the Gemini reversal. Helpfulness scores were also ceiling-compressed. Opus assigned a median of $5$ to \CeilOpus{} of \NturnTotal{} turns, compared with \CeilSol{} for Sol. Sol was systematically more lenient on pedagogy, so directional agreement does not imply scale equivalence or human validity. Each judge also shared a model family with one tutor base: Opus with the Claude Sonnet tutor and Sol with the GPT-5.5 tutor. A Sonnet-and-Gemini-only sensitivity removes the Sol overlap; the Opus overlap remains. Finally, Gemini was a low-leakage boundary case. Its conversational tutor's mean per-session leakage fraction was only
$11.3\%$, compared with $43.0\%$ on Sonnet and $76.6\%$ on GPT-5.5. The leakage manipulation was therefore small and nonsignificant ($\delta{=}\DetLeakGeminiD{}$, $p{=}\DetLeakGeminiP{}$), and both judges scored the policies as pedagogical.

\section{Prompt-versus-Structure Ablation}
\label{sec:ablation}
 
We evaluate five additional registered variants against their baselines under Opus only, descriptively and outside the confirmatory J1/J2 tests: two prompt-only revisions of the conversational tutor (a Socratic instruction or a final-answer ban) and three deletions from the routed pedagogical policy (removing the answer-deferral gate, hint cascade, or learner-state tracker). The main result is that the routed structure is not required to earn the Opus judged-pedagogy score: the Socratic prompt-only variant outscores the full routed policy on judged pedagogy (paired difference ${+}0.565$, CI $[{+}0.348, {+}0.817]$, $\delta{=}{+}0.96$, ten of ten replicate pairs) at near-tied helpfulness. Judged helpfulness stays within a $0.25$-point band ($4.70$ to $4.95$) across all seven policies while judged pedagogy spans $2.3$ points, the \S\ref{sec:primary} divergence, reappearing outside the original contrast (Figure~\ref{fig:dissociation}b), and no single deletion collapses the pedagogy separation.
 
The cascade-free variant is the cautionary case: it raises Opus judged pedagogy (${+}0.535$, $\delta{=}{+}0.96$) and helpfulness (${+}0.151$) yet records the lowest next-turn independence in the pedagogical family, so judge-visible quality and process independence can diverge. Three caveats bound the reading: the Socratic prompt was itself written to be Socratic, narrowing but not dissolving the by-construction concern of \S\ref{sec:design-measures}; the final-answer-ban variant defers disclosure but still leaks on $14.6\%$ of its in-window turns per session; and the tracker-free variant alone drops from two model calls to one, a structure-versus-compute confound. We therefore limit the inference to earning the Opus pedagogy score in this setting, not that pedagogical structure is generally unnecessary. The per-policy couplings, all negative, appear in Figure~\ref{fig:forest}.
 
\section{Discussion}
\label{sec:discussion}
 
\noindent\textbf{Measurement interpretation.} A general-purpose helpfulness score should not be the sole signal for ranking tutors. Under Opus, it did not separate policies that the pedagogy rubric separated perfectly (\S\ref{sec:primary}). The descriptive seven-policy ablation showed the same compression of helpfulness relative to pedagogy (\S\ref{sec:ablation}). Across the two prospectively specified judges, the ConvTutor--PedTutor helpfulness ordering reversed on two of three bases (\S\ref{sec:crossmodel}). The pedagogy-targeted rubric retained its direction wherever it detected a difference. These results show unreliability as a \emph{sole} pedagogy signal, not the impossibility of every calibration scheme.
 
\noindent\textbf{Why the signals differ.} Several non-exclusive mechanisms are consistent with the split: helpfulness ceiling compression, the specificity of a targeted rubric, between-judge scale shifts, and same-family judge--tutor overlap. Sol's greater leniency on pedagogy supports the scale-shift explanation, but we establish none of these mechanisms as causal.
 
\noindent\textbf{Evaluation implication.} These findings support pairing a pedagogy-targeted rubric with deterministic process measures. The only result that held across tutor bases without evaluator dependence came from the deterministic detectors. Leakage and independence could be tested as penalties or filters for pedagogical preference pairs \citep{sonkar2024pedagogical}. We ran no fine-tuning, however, so their value for reward training remains an open empirical question.
 
\noindent\textbf{Scope and boundaries.} The study uses two LLM judges without human validation, one weak simulated student, one algebra domain, and ten replicates per condition, limiting generalization and sensitivity to small effects. Near-ceiling carried-context accuracy does not measure durable learning. The pedagogy contrast is a manipulation check, and the leakage--independence association is observational rather than causal.
 
\section{Conclusion}
 
We asked whether a general-purpose helpfulness rubric can distinguish answer-giving from pedagogical guidance. On the primary base under the frozen Opus judge, the policies did not differ significantly in helpfulness but were perfectly rank-separated on pedagogy. In the prospectively specified GPT-5.6~Sol audit, the helpfulness ordering reversed on two of three bases. The pedagogy-targeted rubric retained its direction wherever a difference was detected. The evaluator-independent result was deterministic: a leaky turn was followed by less independent student work on every base. In this controlled setting, general-purpose helpfulness was not a reliable pedagogy signal. Tutor evaluation should pair pedagogy-targeted rubrics with deterministic process measures.


\bibliography{aaai2027}

\clearpage
\appendix
\raggedbottom

\noindent These appendices document the study design, the problem instrument, and the operational definitions behind every measure and statistical estimand reported in the main text. They also state, for each analysis component, whether that component was specified before or after the evidence bearing on it was visible. Appendix~\ref{sec:supp-design} describes the control structure, the session protocol, and the construction of the frozen 19-problem instrument. Appendix~\ref{sec:measures} gives the measurement definitions. Appendix~\ref{sec:status} records the specification status of each analysis component. Appendices~\ref{sec:annex-prompts}--\ref{sec:annex-problems} release the treatment prompts, the judge rubrics, and the full problem inventory verbatim. No new confirmatory claim is introduced here.

\section{Experimental Design and Problem Instrument}
\label{sec:supp-design}

\subsection{The scientific control}
\label{sec:design-control}

\begin{figure*}[t]
    \centering
    \includegraphics[width=\textwidth]{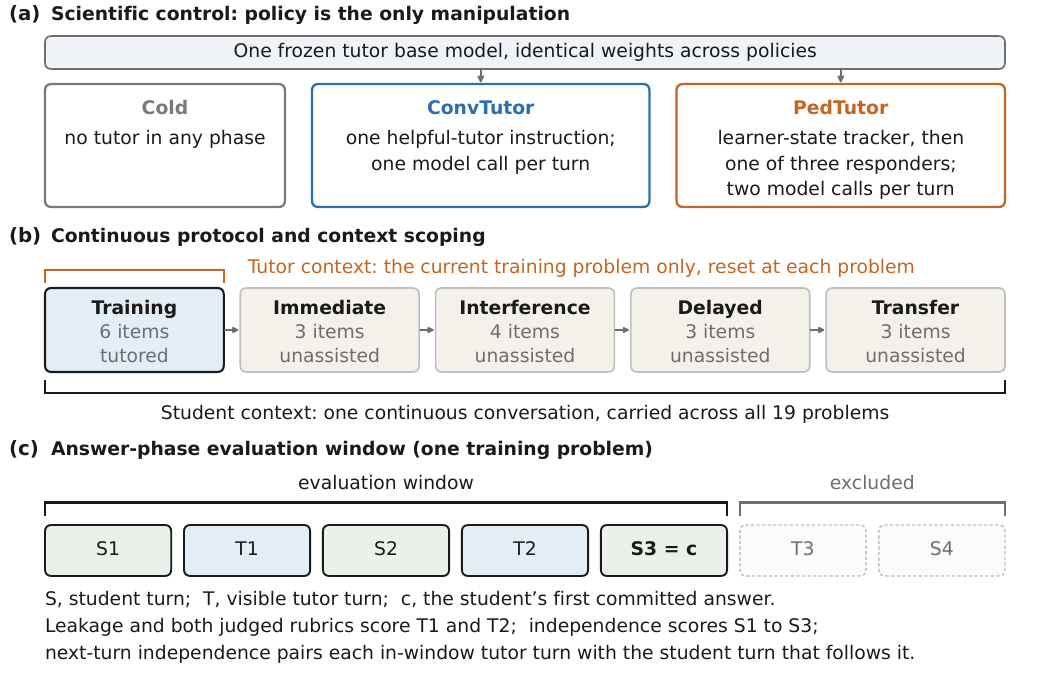}
    \caption{Design, protocol, and measurement window. (a) Both tutoring policies are instantiated from identical frozen weights, so only the policy layer differs; the cold baseline calls no tutor model. All conditions face the same weak simulated student and the same 19 problems. (b) One session is a single continuous protocol, and the two context scopes differ: the student holds one conversation that grows across all 19 problems, whereas the tutor is given only the current training problem's dialogue and sees no probe. (c) Turn-level measures are computed over the answer phase, up to and including the student's first committed answer $c$; later turns are excluded from both conditions by the same rule. Appendix~\ref{sec:measures} defines each measure.}
    \label{fig:supp-design}
\end{figure*}

The study is a within-base comparison of two tutoring policies against a common measurement apparatus. The controlled quantity is the tutor's parameters. Within a tutor base, ConvTutor and PedTutor are instantiated from the same frozen model, served through the same provider under the same sampling configuration, and differ only in the policy layer that sits above the model: the control flow that decides how many calls a visible turn costs, and the instructions those calls carry (Figure~\ref{fig:supp-design}a). No weights are trained, fine-tuned, or adapted at any point in the study, so a difference between the two arms cannot be attributed to a difference in the underlying model.

ConvTutor is the minimal case: a single instruction to be a helpful, clear math tutor, and one model call per visible turn. Its instruction contains no directive about disclosing or withholding answers. Whatever answer-giving it exhibits is therefore an emergent property of a general-purpose helpful assistant applied to tutoring, which is the behavior the study is about, rather than an engineered contrast.

PedTutor is the structured case. Each visible turn begins with a private assessment by the learner-state tracker that never reaches the student, after which a deterministic rule selects exactly one of three responders: one that decomposes the problem into a prerequisite sub-step, one that declines a direct request for the answer and redirects it, and one that issues a hint whose specificity rises with each prior tutor turn on the problem and is held below its most concrete level until the student has made at least two reasoning attempts. Its four nodes correspond to the four tutoring principles cited in Related Work: contingent tutoring to the private assessment, and the assistance dilemma, the generation effect, and tutored help-seeking to the three responders respectively. Two consequences matter for interpretation. First, a visible PedTutor turn costs two model calls rather than one, so the policies are not matched on compute; Appendix~\ref{sec:est-matched} defines the estimand that equalizes visible turns and the separate control that normalizes by tutor tokens. Second, PedTutor is constructed from the same principles that the pedagogy rubric of Appendix~\ref{sec:meas-rubrics} scores, which is why the pedagogy contrast is reported as a manipulation check rather than as independent evidence (\S\ref{sec:design-measures}).

A third condition, cold, runs the identical problem schedule with no tutor in any phase. It calls no tutor model at all and contributes no tutor turns, so it enters the accuracy comparison and the calibration argument but not the turn-level analyses.

The design is replicated on three tutor bases: Claude Sonnet~4.6, GPT-5.5, and Gemini~3.1~Pro Preview, each sampled at temperature $0.4$ under a 1024-token completion limit. Across bases the student, the judges, the problems, the prompts, the evaluation window, and the measures are held fixed. Two things co-vary with the tutor model and are declared rather than controlled. First, each base's tutor is served by that model's own vendor, so the serving provider and the request surface change together with the model; no base is a model swap holding the provider fixed. Second, Gemini's provider enforces a minimum reasoning setting that cannot be switched off, whereas Sonnet and GPT-5.5 run with reasoning disabled. A cross-base difference is therefore not attributable to model family alone, and the Gemini base is the furthest from a pure base-model swap. Neither of these varies within a base, which is where the policy contrast is estimated.

\subsection{Session protocol}
\label{sec:design-protocol}

A session is one uninterrupted episode with a single student instance (Figure~\ref{fig:supp-design}b). It proceeds through five phases in a fixed order: six training problems, then three immediate probes, four interference probes, three delayed probes, and three transfer probes. The training problems are tutored in ConvTutor and PedTutor and untutored in cold; every probe is answered without any tutor, from whatever the student carries in context.

Within a training problem the exchange is a fixed-length alternation. The student is presented with the problem and replies, after which the tutor and the student alternate for four tutor turns, identical in both policies and fixed in the frozen protocol implementation before confirmatory collection; the registered plan fixes the phase structure but not this turn count. The dialogue is not terminated when the student commits an answer, which is why post-resolution turns exist in the log and why the evaluation window of Appendix~\ref{sec:meas-window} is applied after the fact rather than as a stopping rule during collection. On a probe, the student answers once; if that answer carries no parseable final-answer marker, a condition-neutral instruction elicits one, with a single retry. That instruction contains no hint and does not tutor.

The interference block sits between the immediate and delayed probes and is not scored as a learning outcome. Its role is temporal and structural: it separates the delayed probes from training by a block of problems that require a competing solution method, so the delayed probes are not simply a continuation of the training exchange.

Ten replicates are run per condition per base, giving 30 sessions per base and 90 confirmatory sessions in total. A replicate is a pre-registered identifier that fixes the problem order, the condition order, and the student's initial context, and it is shared by the three conditions within that replicate. The replicate, not the individual model call, is the unit of pairing. Because the primary provider exposes no sampling seed, residual generation stochasticity is treated as within-replicate noise and is reported as such rather than eliminated.

\subsection{Student context and tutor context}
\label{sec:design-context}

The two agents see deliberately different amounts of the session, which determines how probe accuracy should be interpreted.

The student holds one conversation for the whole session. Everything presented to it, including each new problem statement and each tutor utterance, is appended to that conversation, and its own replies accumulate alongside. It has no memory beyond that context window and no access to reference solutions. Consequently a probe is not answered from a trained representation but from the transcript of everything that has already happened, including any worked example a tutor supplied earlier.

The tutor holds only the current training problem. Its input is the problem statement together with the dialogue on that problem so far; at the next problem it starts again with nothing carried over. It never sees a probe, never sees another problem's exchange, and is never given a reference answer.

This asymmetry is the reason probe accuracy is a secondary and largely uninformative outcome in this paradigm. A tutor that supplies clean worked examples writes those examples into the student's context, where they persist for the rest of the session. Recoverability from carried context is therefore what the probes measure, and it saturates. The measure cannot separate durable learning from in-context retrieval, because a frozen model has no durable learning to separate. The resulting probe-accuracy ceiling is reported in \S\ref{sec:primary} and treated in \S\ref{sec:design-measures} as a property of the paradigm.

\subsection{The problem instrument}
\label{sec:design-problems}

The instrument is 19 authored problems, frozen before confirmatory collection (Table~\ref{tab:instrument}; the full inventory is Appendix~\ref{sec:annex-problems}). The learning target is a single competence: translating a word problem into a mixture or weighted-average balance equation. Solving that equation once written is linear algebra the student can already do, so the instrument isolates the setup step rather than arithmetic.

\begin{table}[t]
\centering
\small
\setlength{\tabcolsep}{3pt}
\begin{tabular}{@{}l c >{\raggedright\arraybackslash}p{2.05cm} >{\raggedright\arraybackslash}p{3.15cm}@{}}
\toprule
\textbf{Phase} & \textbf{Items} & \textbf{Topic family} & \textbf{Relation to training} \\
\midrule
Training & 6 & mixture & tutored; the only assisted phase \\
\addlinespace[2pt]
Immediate & 3 & mixture & isomorphs of three training items \\
\addlinespace[2pt]
Interference & 4 & relative motion & competing solution method; not scored \\
\addlinespace[2pt]
Delayed & 3 & mixture & isomorphs of three training items \\
\addlinespace[2pt]
Transfer & 3 & price blend, alloy, octane blend & same balance structure, new surface \\
\bottomrule
\end{tabular}
\caption{Structure of the frozen 19-problem instrument. Only training is tutored, and only under the two tutoring policies; the four probe phases are answered without a tutor from the student's accumulated context. Per-item statements, canonical answers, isomorph mappings, and declared leakage forms are in Appendix~\ref{sec:annex-problems}.}
\label{tab:instrument}
\end{table}

\paragraph{Isomorphs.} An immediate or delayed item is an isomorph of a training item when it instantiates the same balance relation with different numbers and, where applicable, a different surface quantity. Isomorphy is a structural claim, not a paraphrase: the equation schema is shared and at least one quantity is changed, so the probe cannot be answered by reproducing the training item's arithmetic; on the three delayed items the only changed quantity is the base amount. Each immediate item maps to one training item, and each delayed item maps to one training item, so a probe failure is attributable to a specific taught relation rather than to the family as a whole.

\paragraph{Interference.} The interference items use relative motion rather than mixture. The choice was made on measured evidence about the student instrument rather than by topical convenience. In calibration the same student could set up motion problems substantially more often than mixture problems, which is precisely the property required: the interfering block must be solvable enough to engage a competing setup schema, rather than merely being an unsolvable delay. Because it uses a different method, it competes with the trained relation in the student's context instead of rehearsing it.

\paragraph{Transfer.} The transfer items retain the weighted-average structure under three new surfaces: blending two priced goods, adding pure material to an alloy, and blending two octane ratings. Each maps to a training item by structure. Transfer here means recognizing the same balance relation when the cover story, the units, and the quantity names all change.

\paragraph{Frozen constraints.} Four constraints on the instrument were verified programmatically before the confirmatory run and are re-verifiable from the released artifact. Every canonical answer is reproduced by an independent symbolic re-derivation written from the problem text rather than copied from the problem file. Every delayed and transfer answer differs from every training answer, so a correct delayed or transfer answer cannot be produced by recalling a training answer; the constraint was not imposed on the immediate phase, where one item shares its answer with a training item. No canonical answer appears as a numeric token anywhere in its own problem statement, so a numeric leakage match cannot be triggered by the problem text itself. Finally, the phase counts match the registered design and every declared isomorph reference resolves.

\subsection{Student calibration and the serving configuration}
\label{sec:design-student}

The simulated student is the one component selected rather than fixed a priori, and the selection criterion was fixed before confirmatory collection. A candidate was accepted only if four conditions held: low accuracy on the frozen problems when working in isolation; failures that inspection of transcripts confirmed to be genuine failures to set up the equation, rather than arithmetic slips or role-played helplessness; a domain demonstrably learnable from tutoring, established by a separate gate in which a tutor lifted untutored performance off the floor; and a parseable final-answer marker emitted on at least 95\% of elicited commitments, without which the accuracy measure is not interpretable.

Screening on isolated accuracy eliminated two candidates before the accepted one. A strong proprietary small model reached ceiling and, on inspection, role-played being stuck while still committing the correct answer, which is the failure mode the simulated-student literature warns about. A 20B open reasoning model likewise reached ceiling. A third serving was rejected on a carried-context reading rather than an isolated one: the same 8B open model that was eventually adopted, served through an fp8-pinned route, answered eight of the nine scored probe items correctly in the calibration pilot for that route. That reading was taken before the isolated and carried-context accuracies were separated (Appendix~\ref{sec:design-context}), so it is reported as the pilot recorded it and is not mode-matched to the isolated figure below; on the isolated criterion that route does not separate from the accepted one.

The accepted instrument is that 8B open model, Llama-3.1-8B-Instruct, served through a single provider route --- reached via OpenRouter and pinned to Groq with fallbacks disabled, so that no substitute provider can be swapped in mid-run --- whose quantization the provider does not report. It is sampled at temperature $0.8$. On that route it solved one of the nine scored probe items when each was attempted in isolation, about $11\%$, in calibration, and none of the nine in the live pre-run re-check; both readings clear the pre-registered low-accuracy ceiling, and transcripts show genuine setup failures. The weak-student precondition is a property of the model and its serving route together, not of the model identifier alone. Reporting the identifier without the route would not reproduce the instrument. The provider pin, the sampling temperature, and the absence of any fallback are therefore part of the specification, and the isolated-accuracy and marker-rate gates, together with the served-provider pin, were re-checked live immediately before the confirmatory run rather than assumed to still hold.

A weak student is a design requirement, not an incidental property. If tutoring is not load-bearing, a tutor that withholds and a tutor that discloses produce the same trajectory, and no process contrast is measurable. The cost of the choice is stated plainly in the scope statement of \S\ref{sec:discussion}: results describe one simulated learner on one algebra domain and do not license claims about human learners.

\section{Operational Measures and Statistical Estimands}
\label{sec:measures}

This appendix defines what each reported quantity is and what it estimates. Notation is fixed as follows. A session is indexed by condition $k$ and replicate $r$. Within a session, $p$ indexes a training problem and $s(\cdot)$ gives a turn's position in the session's call sequence. A \emph{visible tutor turn} is what the student actually received on turn $t$ of problem $p$: for ConvTutor the single call, for PedTutor the responder call, with the private learner-state assessment excluded because the student never saw it. For a visible turn spanning more than one call, $s(t)$ denotes its first constituent call; Appendix~\ref{sec:meas-independence} is the one place where the last constituent call is used instead, and it says so. Turn identity is $(k,r,p,t)$ and is shared by every turn-level measure, so leakage, both judged scores, and next-turn independence align one-to-one on the same turn. Every within-session set and every per-turn quantity below carries the condition index, because the two policies produce different turns on the same problem; the one exception is the matched budget of Appendix~\ref{sec:est-matched}, which is by construction a function of both conditions and therefore carries none.

The five pre-registered claims are labeled as in \S\ref{sec:design-inference}; their operational content is restated here. \textbf{P1}, \textbf{P2}, and \textbf{P3} are session-level policy contrasts on answer leakage, judged helpfulness, and the independence ratio, with registered directions ConvTutor${>}$PedTutor, ConvTutor${>}$PedTutor, and PedTutor${>}$ConvTutor respectively; P2 is the live test and P1 and P3 are manipulation checks. \textbf{J1} is their conjunction, supported only if all three contrasts are significant in the pre-registered direction; P2, the live test, is the component the claim turns on. \textbf{J2} is the turn-level coupling, and it has two legs estimated on the same turns: a turn that discloses the answer is rated more helpful, and it is followed by less independent student work. Every claim is decided at $\alpha = .05$ by a two-sided test together with an observed sign matching the pre-registered direction. For P1, P2, P3 and hence J1 that test is the session-level signed-rank procedure of Appendix~\ref{sec:est-paired}; for J2 it is the two-sided test of $\beta_L$ in \eqref{eq:j2-pooled}, and J2 is supported only when both legs are significant with the registered signs.

\subsection{The answer-phase evaluation window}
\label{sec:meas-window}

Every turn-level measure is computed over the \emph{answer phase} of a training problem, whose inclusive endpoint is the student's first committed answer (Figure~\ref{fig:supp-design}c). Let
\[
  c_{krp} \;=\; \min\bigl\{\, s(u) \;:\; u \in \mathcal{U}_{krp},\ \mathrm{ans}(u) \neq \varnothing \,\bigr\},
\]
where $\mathcal{U}_{krp}$ is the set of student turns on problem $p$ of the session $(k,r)$ and $\mathrm{ans}(\cdot)$ is the frozen final-answer extractor. If no student turn on $p$ yields a parseable answer, $c_{krp} = \infty$. Writing $\mathcal{V}_{krp}$ for the visible tutor turns on the same problem, the in-window sets are
\begin{align*}
  \mathcal{T}_{krp} &= \{\, t \in \mathcal{V}_{krp} \;:\; s(t) < c_{krp} \,\},\\
  \mathcal{S}_{krp} &= \{\, u \in \mathcal{U}_{krp} \;:\; s(u) \le c_{krp} \,\},
\end{align*}
for visible tutor turns and student turns respectively. A turn carrying no marker is not a commitment, which makes the window conservative in the sense of retaining more turns rather than fewer. That protection is one-sided: the extractor keys on the phrase wherever it appears and takes the first number after it, so a turn that uses the phrase without committing can close the window early instead.

Three properties of this window matter for interpretation. It is defined by the student's behavior, not the tutor's, so it cannot be set by either policy directly. It is applied identically to both conditions and to all four turn-level measures, including both judged rubrics: a turn outside the window is scored by nothing. And it is asymmetric in its consequences, which is declared rather than corrected. A policy that resolves a problem quickly earns a short window; a policy that scaffolds toward the answer earns a long one. The window therefore compares tutoring conducted while the problem is still open, which is the comparison of interest, but it does not equalize exposure. Appendix~\ref{sec:est-matched} defines the separate estimand that does.

Session-level cost accounting and probe accuracy are not windowed. Cost is the compute actually spent, and accuracy is measured over the scored probe items of Appendix~\ref{sec:design-protocol}, which have no tutor turns. A probe response counts as correct when the value carried by its final-answer marker matches the item's canonical answer to within $10^{-6}$ in absolute value.

\subsection{Answer leakage and its detection boundary}
\label{sec:meas-leakage}

Each problem $p$ carries two sets declared in advance: a set $N_p$ of numeric forms of its canonical answer, and a set $\Phi_p$ of solution phrasings. Writing $\tilde{x}$ for the normalized text of a visible tutor turn --- case-folded, then reduced to digits, letters, whitespace and the five retained characters \texttt{.}, \texttt{+}, \texttt{-}, \texttt{/} and \texttt{=}, every other character being replaced by a space before whitespace is collapsed --- and normalizing each declared phrase the same way, leakage is the indicator
\[
\begin{aligned}
  L_{krpt} \;=\; \mathbf{1}\bigl[\ &\exists\, n \in N_p:\ n \sqsubset_{\mathrm{tok}} \tilde{x}_{krpt}\\[-1pt]
  \vee\ \ &\exists\, \phi \in \Phi_p:\ \phi \sqsubset \tilde{x}_{krpt}\ \bigr],
\end{aligned}
\]
where $\sqsubset_{\mathrm{tok}}$ denotes a match on token boundaries and $\sqsubset$ ordinary substring containment. The boundary condition on numeric matching is what prevents an answer of $12$ from being detected inside $120$ or inside a decimal; without it the measure would report spurious disclosure whenever the answer's digits occurred incidentally.

What the detector misses and what it may over-call are asymmetric, and only one side is established. Missed reveals are structural and certain: a turn that reveals the answer by a route not enumerated in $N_p$ or $\Phi_p$, most importantly an unlisted paraphrase of the worked solution, is not flagged. Spurious flags are controlled by construction rather than measured. The declared forms were frozen before confirmatory data collection --- they were last revised during calibration, to add decimal variants and to replace digit-bearing solution phrases with spelled-out ones --- and the instrument was built so that no canonical answer appears as a numeric token in its own problem statement, so a numeric hit cannot be produced by the problem text itself. An incidental coincidence, in which the answer's value appears in a turn that does not disclose it, remains possible; its rate was not estimated, and no validation of the detector's specificity was performed.

The consequences are therefore bounded rather than exact. Each condition's rate is a lower bound on its true disclosure rate to the extent that spurious flags are rare, and the measure should be read as an operational index of explicit disclosure rather than of all disclosure. The policy contrast is a difference in that operational index. It transfers to a difference in total disclosure only under the additional assumption that undetected paraphrase occurs at similar rates under both policies. We do not assume that, and we do not tune the declared forms after confirmatory data collection, which would convert a pre-declared index into a fitted one.

The session-level marginal is the fraction of in-window visible tutor turns that leak, pooled over the session's training problems:
\[
  \Lambda_{kr} \;=\; \frac{\sum_{p} \sum_{t \in \mathcal{T}_{krp}} L_{krpt}}{\sum_{p} \lvert \mathcal{T}_{krp} \rvert}.
\]
Because the denominator pools turns rather than averaging problem-level rates, a problem with more in-window turns contributes proportionally more, which matches the quantity of interest: the chance that an arbitrary tutoring turn a student receives discloses the answer.

\subsection{Student independence at two levels}
\label{sec:meas-independence}

The primitive is a per-turn indicator on student turns. A student turn attempts a reasoning step when it contains explicit symbolic or numeric work: an equation relating two expressions, an arithmetic operation between two numbers, a coefficient attached to a variable, or a variable standing in an operator relation. Writing $\mathcal{W}$ for the set of student turns exhibiting at least one of these four forms, the primitive is
\[
  A_{krpu} \;=\; \mathbf{1}\bigl[\, u \in \mathcal{W} \,\bigr] \in \{0,1\}.
\]
A bare number and a bare answer guess in prose do not qualify, because neither exhibits a reasoning step; a guess written symbolically, as an equation with the unknown on one side, does satisfy the first form and is counted. This choice is conservative with respect to the pre-registered direction: the hypothesis predicted that PedTutor would elicit more independent work, so a definition that credited unsupported guesses would inflate exactly the quantity under test. Paraphrased reasoning carrying no symbolic content is likewise not captured, so the measure detects explicit work and under-counts implicit work in both conditions.

This single primitive supports two distinct estimands, and the distinction is the reason they are reported separately in \S\ref{sec:design-measures}.

\paragraph{Session-level independence ratio (P3).} The fraction of a session's in-window student training turns that attempt a step,
\[
  I_{kr} \;=\; \frac{\sum_{p} \sum_{u \in \mathcal{S}_{krp}} A_{krpu}}{\sum_{p} \lvert \mathcal{S}_{krp} \rvert},
\]
a session summary that is compared between policies, paired by replicate. Its unit is the session; there are ten per condition per base.

\paragraph{Turn-level next-turn independence (J2).} For a visible tutor turn $t$, let $\nu(t)$ be the first student turn strictly after the last constituent call of $t$ on the same problem. The outcome is
\[
  Y^{\mathrm{ind}}_{krpt} \;=\; A_{krp\,\nu(t)} \in \{0,1\},
\]
undefined when no such student turn exists, in which case the tutor turn is dropped from that analysis. Under the fixed alternation of Appendix~\ref{sec:design-protocol} every in-window tutor turn is followed by a student turn, so no turn is dropped on this ground in any confirmatory corpus and both J2 legs are fitted on the same turns. This quantity is compared \emph{between leaky and non-leaky turns}, pooled across both policies, within a base.

The two therefore answer different questions from the same labels. $I_{kr}$ asks whether one policy leaves the student doing more of the work over a session. $Y^{\mathrm{ind}}_{krpt}$ asks whether a turn that discloses the answer is followed by less student work, regardless of which policy produced it. They can move independently, and in the reported data they do: a session-level contrast can fail to separate while the turn-level coupling is strongly detected, because the latter has far more units and conditions on a within-session event rather than on policy assignment.

Both labels, and the leakage labels of Appendix~\ref{sec:meas-leakage}, are produced by fixed rules applied to the transcript text. No model relabels them. This is what makes both deterministic outcomes judge-invariant by construction, and it is why Sol could not rescore them: there is nothing for a judge to score. The registered metric definition has the rule-based label confirmed by a model-based verification pass, with the rule-based label authoritative and disagreements logged rather than applied. That pass was implemented as opt-in and was not exercised on the confirmatory corpora, whose independence values therefore contain no model adjudication of any kind; the reported labels are the unconfirmed rule-based ones, which is a deviation from the registered definition. Its rubric is released in Appendix~\ref{sec:annex-rubrics} because it defines the intended reading of the primitive.

\subsection{What the two rubrics measure}
\label{sec:meas-rubrics}

Two rubrics are applied to the same in-window visible tutor turns by the same judge, with the same repetition scheme and the same context. They target different constructs, and the divergence between them is the study's central quantity (\S\ref{sec:primary}), so the constructs are worth separating precisely.

\paragraph{Annotator-perceived helpfulness.} This instrument estimates how helpful a turn appears to an external evaluator applying a general-purpose rubric: whether the turn is clear, whether it engages with what the student actually said, and whether it gives the student something to move forward with in the moment. It is a reward-model-style proxy for an annotator preference signal. Two exclusions define its scope, and both are stated in the instrument itself. It does not evaluate teaching strategy, and it explicitly instructs the rater not to reward or penalize disclosing or withholding an answer. It does not evaluate mathematical correctness. It therefore measures neither the learner's experience nor pedagogical quality, and the strategy exclusion is what makes it a fair test: an instrument that already encoded a preference about answer-giving could not be used to ask whether general-purpose helpfulness detects answer-giving.

\paragraph{Judged pedagogy.} This instrument estimates the pedagogical quality of the same move on four principles: whether support is contingent on the student's diagnosed difficulty, whether the reasoning the student could still generate is preserved, whether the amount of assistance is calibrated between floundering and doing the work for them, and whether the turn elicits and builds on the student's own contribution. It is written symmetrically. Disclosure is not penalized categorically and withholding is not rewarded categorically; a well-targeted hint that reveals part of the answer can score highly, and a generic exhortation to keep trying that ignores the student's stated confusion can score poorly. The rubric nonetheless scores the principles that PedTutor was built from, which is the construction dependence declared in \S\ref{sec:design-measures} when it reports the pedagogy contrast as a manipulation check.

Each rubric yields component scores and one holistic rating. The per-turn metric is the holistic rating in both cases; component scores are retained for description and are not aggregated into the metric.

\subsection{The judge context}
\label{sec:meas-judgecontext}

What a judge is shown determines what its score can mean, so the context is specified exactly. For a rated turn $(k,r,p,t)$ the judge receives the rubric and one reconstructed dialogue $D_{krpt}$, and nothing else. $D_{krpt}$ contains the student's turns and the visible tutor turns on problem $p$, each restricted to $s(\cdot) \le s(t)$, in sequence order, ending with the turn under evaluation. No student turn after the rated turn enters it, so a judged score cannot be informed by the behavior that next-turn independence records. It is assembled from logged turn text only.

Four things are never injected into that input from the run record: the problem's canonical answer, the declared answer forms of Appendix~\ref{sec:meas-leakage}, the internal node labels that would identify which responder produced a PedTutor turn, and the condition label. A judge therefore cannot look up the answer in the record, is never shown the detector's string lists, and is given no label naming the policy.

The problem statement is presented to the student as a framing instruction rather than as a logged conversational turn and is therefore not reproduced verbatim in $D_{krpt}$. It is nonetheless often recoverable from the exchange, because the student is instructed to restate the problem sometimes and the tutor, which has access to the statement, may quote it. Information about the problem therefore reaches the judge only through the dialogue between the student and the tutor.

This does not make the judged scores independent of the leakage labels by construction. A turn leaks by containing the answer in its text, and that text is exactly what the judge reads. What holds instead is weaker and sufficient. The judge is never given the leakage label, is instructed not to verify the mathematics, and applies a helpfulness rubric that explicitly excludes disclosure strategy from the score; the pedagogy rubric weighs disclosure case by case instead, so only the first two protections carry over to it. Any association between leakage and a judged score is consequently a property of how the rubric responds to a turn's content, not a definitional link imposed by the measurement pipeline.

One limitation follows directly. Because neither rubric checks the mathematics against ground truth, neither can detect a turn that is fluent, responsive, and wrong about this particular problem. The paper's claims are correspondingly about what a general-purpose helpfulness signal rewards, not about tutor correctness.

Both judges receive identical context for a given turn, because the reconstruction, the rubrics, the repetition count, the parsing, and the aggregation rule are shared. Their configurations are not otherwise identical, and Appendix~\ref{sec:est-crossjudge} declares what else differs. The frozen primary judge is Claude Opus~4.8 and the post hoc robustness judge is GPT-5.6~Sol.

\subsection{Aggregation}
\label{sec:meas-aggregation}

Aggregation proceeds through three levels and never skips one.

Each turn is rated three times under each rubric. The aggregation defined here is applied separately for each instrument $m \in \{\text{helpfulness}, \text{pedagogy}\}$ and each judge $q$; both indices are suppressed below and restored as $S^{(m)}_{krpt,q}$ wherever a comparison ranges over them. The per-turn score is the mean of the valid holistic ratings,
\[
  S_{krpt} \;=\; \frac{1}{\lvert \mathcal{R}_{krpt} \rvert} \sum_{j \in \mathcal{R}_{krpt}} h^{(j)}_{krpt},
\]
where $\mathcal{R}_{krpt}$ indexes the repetitions that produced a usable holistic rating. A repetition is usable when a holistic rating can be read from its response --- from the widest brace-delimited span in it when that span parses as JSON, with a per-field pattern match filling every field the parse leaves unset --- and that value rounds to an integer in $1$--$5$; a response from which no holistic rating can be read is discarded and counted rather than imputed. Repetition spread is summarized by the sample variance across those ratings, with Bessel's correction, and is undefined with fewer than two usable ratings. Because the primary judge's sampling parameters are left at provider default rather than pinned, the three repetitions are genuine stochastic samples and their spread is real instrument variability, not a synthetic perturbation.

The per-session score is the unweighted mean of the per-turn scores over the session's scored in-window visible tutor turns. Writing $\mathcal{T}^{+}_{krp} = \{\, t \in \mathcal{T}_{krp} : \mathcal{R}_{krpt} \neq \varnothing \,\}$ for those with at least one usable rating,
\[
  \bar{S}_{kr} \;=\; \frac{1}{\sum_{p}\lvert \mathcal{T}^{+}_{krp}\rvert} \sum_{p}\sum_{t \in \mathcal{T}^{+}_{krp}} S_{krpt}.
\]
A turn for which no repetition produced a usable rating is dropped from the session mean rather than entered as a zero. In the confirmatory corpora no turn was dropped on this ground, so the two denominators coincide there. The per-replicate value is the mean over sessions sharing $(k,r)$. With one session per condition per replicate, the per-replicate value equals the session value.

Pooled turn-level averages are never used for inference. Turns within a session are correlated, so treating them as independent units would overstate the effective sample size. Every session-level test uses ten paired units per base (\S\ref{sec:design-inference}), and every turn-level analysis carries an explicit dependence structure (Appendix~\ref{sec:est-crossed}). That structure is specified at the replicate and problem level; a replicate spans both policy sessions, so within-session correlation is absorbed only to the extent that it is shared across the pair.

\subsection{Paired inference on session marginals}
\label{sec:est-paired}

For a session-level measure $M$, the estimand is the mean paired policy difference within a base,
\[
  \Delta_M \;=\; \mathbb{E}\bigl[\, M_{\mathrm{conv},r} - M_{\mathrm{ped},r} \,\bigr],
\]
with the expectation over replicates. Pairing is by replicate, which is the pre-registered unit and which holds the problem order, the condition order, and the student's initial context fixed across the two arms of a pair.

Inference uses the two-sided Wilcoxon signed-rank test on the ten paired differences at $\alpha = .05$, with zero differences excluded from the statistic. The implementation passes no method argument and lets the library select rather than forcing either its exact or its asymptotic form; at this sample size that selection never reaches the tie-corrected normal approximation, so the reported session-level $p$-values are exact rather than asymptotic. Where the absolute differences carry ties the selection is additionally version-dependent, and the primary base was computed under a newer library version than the two cross-base replications. On the one contrast where the two selections differ, the leakage contrast on the Gemini base, the older version returns the tie-free reference value $.625$ and the newer one the tie-corrected permutation value $.623$; the released inference artifact for that base records the tie-free value and the decision log records both, and no reported decision turns on the difference. Effect size is reported as Cliff's $\delta$ computed between the two unpaired groups of session values, which is why $\delta$ can differ in sign from $\Delta_M$ when the pairing carries the contrast. Interval estimates for $\Delta_M$ are percentile bootstrap 95\% intervals over the paired differences. The point estimate and its interval therefore target $\Delta_M$ itself, whereas the signed-rank statistic targets the center of symmetry of the paired-difference distribution, its Hodges--Lehmann pseudomedian. The two functionals coincide when the differences are symmetric about a common center and can otherwise diverge, so a $p$-value and an interval that disagree at the margin are not in contradiction. We report both rather than substituting one for the other.

A pre-registered claim counts as supported only under the conjunction of a two-sided $p < \alpha$ and an observed sign matching the pre-registered direction. A two-sided test with a directional decision rule is deliberate: it does not allow a significant reversal to be counted as support, and it forces a reversal to be reported as such. With ten paired units, the session-level estimates remain imprecise. We therefore report effect sizes and confidence intervals and do not interpret nonsignificance as equivalence or as excluding effects below a prespecified threshold.

\subsection{Crossed-effects turn-level models}
\label{sec:est-crossed}

The turn-level analysis asks whether a turn that discloses the answer differs from one that does not, on two outcomes: the judged helpfulness of that turn, and whether the following student turn attempts a reasoning step. Turns are nested in neither replicates nor problems exclusively, since every replicate encounters every training problem, so the dependence is crossed rather than hierarchical. The registered specification is
\begin{equation}
  Y_{krpt} \;=\; \beta_0 + \beta_L L_{krpt} + u_r + v_p + \varepsilon_{krpt},
  \label{eq:j2-pooled}
\end{equation}
with $u_r \sim \mathcal{N}(0,\tau^2_r)$ over replicates, $v_p \sim \mathcal{N}(0,\tau^2_p)$ over problems, and $\varepsilon_{krpt} \sim \mathcal{N}(0,\sigma^2)$ for the continuous judged outcome, fitted by maximum likelihood separately within each tutor base and, for the judged outcome, separately by judge. The post hoc sensitivity specification adds tutoring policy as a fixed effect,
\begin{equation}
  Y_{krpt} \;=\; \beta_0 + \beta_L L_{krpt} + \beta_C C_k + u_r + v_p + \varepsilon_{krpt},
  \label{eq:j2-adjusted}
\end{equation}
with $C_k = 1$ when $k$ is PedTutor and $0$ when it is ConvTutor. It supplements rather than replaces \eqref{eq:j2-pooled}.

The two coefficients estimate different things, and conflating them would misread the analysis. In \eqref{eq:j2-pooled}, $\beta_L$ pools within-policy and between-policy variation: because leakage prevalence differs sharply by policy, part of that coefficient reflects the policy contrast itself. In \eqref{eq:j2-adjusted}, $\beta_L$ is the leakage contrast holding policy fixed, that is, the difference between leaky and non-leaky turns produced by the same policy. Where the two disagree, the disagreement localizes the association to the between-policy component and is reported as such.

For the binary independence outcome a $\{0,1\}$ response admits only two residual values at any linear predictor, so the Gaussian terms of \eqref{eq:j2-pooled} and \eqref{eq:j2-adjusted} are working likelihoods rather than distributional claims and both fits are linear probability models: $\beta_L$ is a risk difference on the probability scale rather than a log-odds ratio, and the reported standard errors are model-based, uncorrected for the heteroskedasticity a binary response induces. A logistic mixed model is a defensible alternative that was deliberately not substituted after the fact, because the registered specification named a mixed-effects model without a family and the linear family matches the helpfulness leg, keeping the two legs on a common specification.

Leakage is not randomized at the turn level. It is a property of what the tutor chose to say, inside a policy the session was assigned to at the outset. Both coefficients are therefore observational associations between a turn's content and its consequences, not causal effects of disclosure. Part of any drop in subsequent student work is mechanical: once the answer sits in the student's context, there is less for the student to derive. A two-stage clustered procedure, in which a leaky-minus-non-leaky difference is formed within each replicate and those differences are tested across replicates, is reported alongside as a dependence-light sensitivity analysis. Its point estimate is the unweighted mean of those replicate-level differences and its percentile bootstrap interval resamples them, whereas its signed-rank $p$-value targets their Hodges--Lehmann pseudomedian; as in Appendix~\ref{sec:est-paired}, the two functionals can diverge, so an interval and a $p$-value that disagree at the margin are not in contradiction. It is not part of the registered specification: it was written during analysis, after the confirmatory data existed, as a dependency-light substitute when the registered mixed model could not initially be fitted in the analysis environment, and it is retained alongside the registered fit rather than in place of it. It makes weaker distributional assumptions, but its unit is the replicate, which spans both policy sessions: the leaky and the non-leaky turns entering a replicate's difference are drawn from both arms. It is therefore no more policy-controlled than \eqref{eq:j2-pooled}, and it is not a within-session comparison. Because leakage prevalence differs sharply by policy, in several replicates every leaky turn comes from ConvTutor.

\subsection{The matched visible-turn budget}
\label{sec:est-matched}

The policies do not spend equal numbers of visible turns, so a contrast computed over full answer-phase windows compares policies that differ both in what they say and in how much they say. The matched visible-turn-budget analysis targets a different estimand: the policy contrast at a common count of visible turns. It equalizes turns only: a PedTutor visible turn still costs two model calls to a ConvTutor turn's one, so the two arms remain unequal in model calls and in tutor tokens under this budget. The separate control defined at the end of this section normalizes by tokens instead.

For each problem--replicate cell, define
\[
  K_{rp} \;=\; \min\bigl(\lvert \mathcal{T}_{\mathrm{conv},rp}\rvert,\ \lvert \mathcal{T}_{\mathrm{ped},rp}\rvert\bigr),
\]
the pairwise minimum of the two conditions' in-window visible tutor-turn counts. Both conditions are truncated to their first $K_{rp}$ visible tutor turns, together with the student turns preceding the first dropped tutor turn. A cell with $K_{rp} = 0$ is dropped from both conditions. The three count- and score-based session marginals --- answer leakage, judged helpfulness, and the independence ratio --- and both turn-level legs are then recomputed on the truncated transcripts under the same aggregation and the same inference. The judged-pedagogy marginal is not recomputed, so the helpfulness--pedagogy divergence is not evaluated at a common visible-turn budget.

Three properties define the estimand. The rule is symmetric: it is a function of both conditions' turn counts and cannot favor either. It degenerates exactly to the full window when a condition's own count is already the minimum and positive; at $K_{rp} = 0$ the cell is removed instead, so the condition whose count was the minimum loses its in-window student turns as well. And it is a single rule, fixed before any truncated statistic was computed, with no alternative budget explored.

The estimand is not a bias correction on the full-window contrast; it is a distinct comparison, and the two can legitimately differ. Its cost is power and coverage, and on the primary base both are realized. Truncation removes cells entirely where one condition had no in-window turns: five of the sixty problem--replicate cells are dropped at $K_{rp} = 0$, in every case because ConvTutor committed an answer before any in-window tutor turn. Of that base's 358 answer-phase turns, 256 survive. The leaky turns that identify the turn-level legs fall from 52 to 51 under ConvTutor but from 14 to 6 under PedTutor. Six events is below the floor of ten leaky turns per condition that the sensitivity analysis uses as its threshold for an adequate cell. That floor is an analysis-time choice made when the sensitivity was implemented, not a figure carried in the registered plan, which requires only that the cell sizes be reported and that an inadequate cell be described as underpowered rather than repaired by re-specification. Where the floor is not met the leg is reported as resting mainly on the other condition's turns rather than as a balanced within-policy comparison. A separate registered compute-normalized control expresses the two count-based marginals per unit of tutor-token expenditure. Writing $\Gamma_{kr}$ for the total tokens over all of the session's tutor calls,
\begin{align*}
  \Lambda^{\mathrm{tok}}_{kr} &= \frac{10^{3}\sum_{p}\sum_{t \in \mathcal{T}_{krp}} L_{krpt}}{\Gamma_{kr}},\\[2pt]
  I^{\mathrm{tok}}_{kr} &= \frac{10^{3}\sum_{p}\sum_{u \in \mathcal{S}_{krp}} A_{krpu}}{\Gamma_{kr}},
\end{align*}
counts of leaky tutor turns and of independent student turns per thousand tutor tokens. This analysis targets compute-normalized rates rather than equal exposure in visible turns. The two scopes differ by construction: the numerators are in-window counts while $\Gamma_{kr}$ is unwindowed session cost (Appendix~\ref{sec:meas-window}), so these rates are not a windowed quantity divided by a windowed one. Its results are released with the accompanying artifacts and are not carried into the paper's verdicts. Four of its six contrasts reach significance, all in the pre-registered direction except the independence marginal on the Gemini base, where normalization turns a null unnormalized contrast into a significant one favoring ConvTutor ($p = .002$). The released automatic flag compares significance in the pre-registered direction before and after normalization, so a reversal of that kind does not raise it.

\subsection{The cross-judge difference estimand}
\label{sec:est-crossjudge}

Sol rescored the identical set of answer-phase tutor turns ($\NturnSonnet{}$, $\NturnGPT{}$, and $\NturnGemini{}$ turns on the three bases, $\NturnTotal{}$ in total, yielding $\NratingTotal{}$ ratings across the two rubrics) under the same rubrics, the same reconstructed dialogues, the same repetition count, the same parser, the same aggregation rule, the same answer-phase window, and the same condition blinding.

Four aspects changed alongside the judge. First, Sol used a 2048-token completion cap rather than Opus's 512-token cap. Across the Sol corpus, the largest billed completion was 536 tokens, including 501 reasoning tokens. The largest visible completion was 44 tokens, and every Sol record ended with \texttt{stop}. Second, medium reasoning was requested but was not independently attested by the route. Third, Sol repetitions were seeded by repetition index, whereas Opus repetitions were unseeded. Fourth, Sol was served through an intermediary pinned to the OpenAI upstream with fallbacks disabled; every retained Sol record names OpenAI as the serving upstream. All frozen Opus and Sol turns yielded three valid holistic ratings, but the retained Opus artifacts do not preserve finish reasons and therefore do not support a transport-level truncation claim.

Per-turn aggregates are aligned by exact turn identity, so the comparison is within-turn rather than between distributions. For instrument $m \in \{\text{helpfulness}, \text{pedagogy}\}$ the estimand is the within-turn judge difference
\[
  D^{(m)}_{krpt} \;=\; S^{(m)}_{krpt,\,\mathrm{Sol}} \;-\; S^{(m)}_{krpt,\,\mathrm{Opus}},
\]
and fitting \eqref{eq:j2-adjusted} to $D^{(m)}$ yields the judge-by-leakage and judge-by-policy interaction terms directly, because both judges' scores rest on the same turns and therefore on the same design matrix. Multiplicity is controlled by Holm correction applied separately within three families of three base-wise tests, all of them on the helpfulness instrument: Sol's policy-adjusted leakage slopes, the unstandardized judge-by-leakage terms from $D^{(m)}$, and the unstandardized judge-by-policy terms from $D^{(m)}$. The corresponding pedagogy terms and the standardized fits are reported without correction.

Two constraints on the design bound what it can support. The two judges are reported separately and are never averaged into a consensus score: disagreement between evaluators is the quantity of interest, and averaging would destroy it. And $D^{(m)}$ is defined only for judged scores, so the deterministic outcomes have no cross-judge analogue; they are invariant across judges because no judge produces them. A second evaluator bounds robustness across the two specified judges. It does not establish agreement with human annotators, which remains a separate study.

\begin{table*}[t]
\centering
\small
\setlength{\tabcolsep}{4pt}
\begin{tabular}{@{}>{\raggedright\arraybackslash}p{3.05cm} >{\raggedright\arraybackslash}p{3.70cm} >{\raggedright\arraybackslash}p{4.55cm} >{\raggedright\arraybackslash}p{4.75cm}@{}}
\toprule
\textbf{Analysis component} & \textbf{Specification status} & \textbf{Evidence visible when specified} & \textbf{Role in the paper} \\
\midrule
Accuracy-primary hypothesis & Initial pre-registration & Calibration only; no confirmatory data & Abandoned; the ceiling is reported as a paradigm limitation \\
\addlinespace[2pt]
Process outcomes (leakage, judged helpfulness, independence) & Amendment, before confirmatory collection & Calibration only & Confirmatory; judged helpfulness is the live test, the two deterministic marginals are manipulation checks \\
\addlinespace[2pt]
Answer-phase evaluation window & Amendment, before confirmatory collection & Non-inferential pilot only & Confirmatory; the measurement window for every turn-level outcome \\
\addlinespace[2pt]
Cross-base replications & Registered additive extension & Primary-base results & Confirmatory, re-run per base under the frozen inference \\
\addlinespace[2pt]
Pedagogy rubric & Registered after the primary study, frozen before any pedagogy scoring & Frozen transcripts and frozen primary-judge helpfulness scores & Manipulation check; supplies the second rubric of the central divergence \\
\addlinespace[2pt]
Matched visible-turn budget & Sensitivity analysis; rule fixed after the full-window result was read, before any truncated statistic was computed & Full-window primary results & Robustness under a common visible-turn budget \\
\addlinespace[2pt]
Cross-judge audit & Prospectively specified post hoc, with claim limits fixed in advance & Frozen transcripts and frozen primary-judge scores & Judge robustness across Opus and Sol \\
\addlinespace[2pt]
Prompt-versus-structure ablations & Registered descriptive extension & Primary-base and cross-base results & Mechanistic evidence; run on the primary base only and scored by the primary judge only; never enters a confirmatory verdict \\
\bottomrule
\end{tabular}
\caption{Specification status of each analysis component. ``Evidence visible when specified'' is the evidence that existed at the moment the component was fixed, which is what determines whether the component could have been shaped by the result it reports. A component carries confirmatory weight when it was specified before the data it is computed on existed; on that criterion the cross-base replications are confirmatory for their own bases even though the primary-base results were already visible. Components specified after the data they report on are manipulation checks, robustness analyses, or descriptive extensions, and no such component enters a pre-registered verdict.}
\label{tab:status}
\end{table*}

\section{Analysis Status and Deviations from the Initial Plan}
\label{sec:status}

Table~\ref{tab:status} records, for each component that carries a claim in the paper, when it was specified, what evidence was already visible at that moment, and what role it plays in the argument. The distinction the table is designed to make is between claims that were genuinely tested against evidence not yet seen and claims that are explanatory extensions built on results already in hand. Both kinds appear in the paper; only the first kind carries confirmatory weight.

The primary outcome was changed, and the change was substantive: the study began with a durable-learning accuracy hypothesis and abandoned it when the learnability gate lifted immediate, delayed and transfer accuracy under ConvTutor from near zero to ceiling on all three, leaving no headroom for the dissociation the hypothesis predicted --- the saturation that Appendix~\ref{sec:design-context} explains, after which the primary claim was tightened again to the joint process coupling actually tested. Both steps were taken before any confirmatory data existed, which is what makes them legitimate, and the abandoned hypothesis is recorded here rather than removed. The matched visible-turn-budget rule was fixed after the full-window result had been read but before any truncated statistic was computed; it is a robustness analysis specified under partial knowledge, and the ordering is stated precisely rather than presented as fully blind. Sol was named, together with its claim limits, before it scored anything, but it necessarily operated on transcripts and primary-judge scores that already existed, so it is prospectively specified and post hoc at the same time.

Four further components carry a specification status that is stated where each is defined rather than in the table, because each attaches to a particular estimand or measurement definition rather than to a stage of the argument. The per-tutor-token control of Appendix~\ref{sec:est-matched} is registered, and its results are released rather than carried into any verdict. The two-stage clustered procedure of Appendix~\ref{sec:est-crossed} is not registered: it was written during analysis, after the confirmatory data existed. The policy-adjusted specification of Equation~\eqref{eq:j2-adjusted} is post hoc and supplements the registered fit rather than replacing it. And the model-based confirmation pass that the registered independence definition calls for was not exercised on the confirmatory corpora, which Appendix~\ref{sec:meas-independence} records as a deviation from that definition; it is the one deviation in the study that was not a change of plan made in advance.

The reporting rule that governs all of this was fixed with the analysis plan: every registered claim is reported regardless of outcome, a manipulation check that fails to separate is recorded as inconclusive rather than as a falsification, and a live test that is null or reversed is recorded as not supported. No analysis was re-specified to move a $p$-value; concerns about a specification are disclosed alongside the registered fit rather than resolved by replacing it.

\section{Treatment Prompts}
\label{sec:annex-prompts}

The prompts below define the intervention. They are reproduced verbatim from the frozen configuration. The problem statement is appended to each tutor prompt at run time, identically in both policies; no canonical answer appears in any prompt.

\subsection{Conversational policy (ConvTutor)}

One instruction, one model call per visible turn.

\begin{lstlisting}[style=instrument]
You are a math tutor helping a student. Be helpful and clear.
\end{lstlisting}

\subsection{Student simulator}

\begin{lstlisting}[style=instrument]
You are a student learning algebra. You know:
- Basic arithmetic.
- How to solve a linear equation once it is set up.
You do NOT know:
- How to translate a word problem into an equation.
- Which quantity to assign to a variable.
- How to identify the underlying relationship in rate/work/mixture problems.

Behave like a real, somewhat-struggling student. Translating the words into an
equation is the hard part for you; your arithmetic is usually fine once an
equation is set up.
- When you genuinely cannot see how to set the problem up, you sometimes ask for
  help or just ask for the answer.
- Sometimes you restate the problem instead of making progress.
- Otherwise you attempt a step.
Do not invent a method you are unsure of. Keep replies short (1-3 sentences). When
you are ready to commit to a final answer, write it on its own line as:
FINAL ANSWER: <number>
If you are explicitly asked to state your final answer, you MUST reply with that
exact line and nothing else.
\end{lstlisting}

Each problem is presented to the student as an opening message rather than as part of this system prompt, and every tutor utterance is appended to the same conversation.

\subsection{Forced commitment}

Applied identically in all three conditions at every probe item when the student has not already committed. A tutored training item is never elicited; in the cold condition, where training is untutored, training items follow the probe path and are elicited in the same way. It contains no hint and does not tutor.

\begin{lstlisting}[style=instrument]
You must now commit to your best single numeric guess for the ORIGINAL problem. Even if you are unsure or would rather ask for help, you MUST write an actual number — do not ask a question and do not leave it blank. Reply with exactly one line and nothing else:
FINAL ANSWER: <number>
\end{lstlisting}

Retry, issued only when the first commitment carried no number:

\begin{lstlisting}[style=instrument]
That had no number. Write your best guess as a single number now — any number is better than none. One line, nothing else:
FINAL ANSWER: <number>
\end{lstlisting}

\subsection{Structured policy (PedTutor)}

Each visible turn runs the learner-state tracker, then exactly one responder selected by a fixed rule: the deferral gate when the student's last turn requests the answer; otherwise the decomposer when the student has made no reasoning attempt yet; otherwise the hint cascade. Only the assessment fields \texttt{demonstrated}, \texttt{missing}, \texttt{stuck}, and \texttt{progressed} are forwarded to a responder, as private notes.

\subsubsection{Learner-state assessment (private; never shown to the student)}

\begin{lstlisting}[style=instrument]
You are the private planning component of a math tutor. You do NOT talk to
the student and nothing you write is shown to them.

Read the dialogue so far and judge where the student is right now. Reply with
ONE line of compact JSON and nothing else:
{"demonstrated": "<what the student has actually shown they can do, or 'nothing yet'>", "missing": "<the single most important next step they have not taken>", "stuck": <true or false>, "progressed": <true if they did something new since the previous tutor turn, else false>, "asked_for_answer": <true if they are asking you to just give the answer>}

Judge only from what the student has written. Do not solve the problem and do
not include the answer anywhere in your assessment.
\end{lstlisting}

\subsubsection{Deferral gate}

\begin{lstlisting}[style=instrument]
You are a patient math tutor. The student is asking you to give them the
answer or the full worked solution. Do not give it. People learn by
generating the steps themselves and by retrieving them with effort, so your
job here is to keep the student thinking, not to hand over the result.

Always:
- Never state or compute the final numeric answer to the problem.
- Never write out the complete solution or the fully-formed equation for them.
- Be warm: acknowledge what they asked, then turn it back into a next step
  they can take.

Make exactly one short move (2-4 sentences): either ask a single focused
diagnostic question aimed at the step they are missing, or give the smallest
hint that points at the relationship in the problem, in words rather than as
a finished equation.
\end{lstlisting}

One of two runtime directives is appended, according to whether the student has reached the attempt threshold.

\begin{lstlisting}[style=instrument]
The student has not yet made enough attempts of their own. Stay conceptual: ask one focused diagnostic question, or describe the relationship in words. Do not give the equation in finished form, and do not state or compute any number that is the answer.
\end{lstlisting}

\begin{lstlisting}[style=instrument]
The student has now made at least two real attempts. You may name the quantities that have to balance and the general form of the equation, but they must build it and do the arithmetic. Still do not state or compute the final numeric answer.
\end{lstlisting}

\subsubsection{Decomposer}

\begin{lstlisting}[style=instrument]
You are a math tutor who teaches by decomposition. The student is stuck on
the problem as a whole. Do not attempt the whole thing. Pick the single most
useful prerequisite sub-step and ask about just that: a smaller question the
student can actually answer, which moves them toward setting the problem up.

Always:
- Ask about ONE sub-step only (for example, what a stated percentage means
  for an actual amount, or which quantity should be the unknown), not the
  full setup.
- Do not state or compute the final numeric answer, and do not hand over the
  full equation.
- One or two sentences, ending in a question.
\end{lstlisting}

\subsubsection{Hint cascade}

\begin{lstlisting}[style=instrument]
You are a math tutor giving a hint. Give the LEAST specific hint that has a
chance of unblocking the student, and become more specific only when your
earlier hints have not worked. You will be told the current hint level:

- abstract: name the general principle or kind of relationship to think
  about, in words. No equation.
- pointed: point to the specific quantities in THIS problem that have to be
  related, still in words or as a partial expression.
- concrete: you may give the form of the equation to set up, but the student
  must do the algebra and the arithmetic themselves.

Always:
- Never state or compute the final numeric answer.
- Acknowledge any correct progress the student has made.
- Two to four sentences.
\end{lstlisting}

Two further directives are templated. A progress note carried from the learner-state assessment, \texttt{Student made new progress since the last tutor turn: \{progressed\}.}, is available to the responders; the hint cascade additionally receives its current level as \texttt{Current hint level: \{level\} (\{name\}).}

\subsubsection{Routing parameters}

The hint level equals the number of prior tutor turns on the problem, over the three levels \texttt{abstract}, \texttt{pointed}, and \texttt{concrete}, and is capped below \texttt{concrete} until the student has made at least two reasoning attempts. That attempt count and answer-request detection are both computed from fixed, case-insensitive patterns over the student’s turns, making the routing decision auditable and independent of model judgment. The attempt patterns require an operator, an equality, an explicit variable-assignment phrase, or a percentage taken of a quantity; the last of these credits a turn that names a percentage of a stated amount without operating on it. The answer-request patterns were deliberately broadened against phrasings observed in calibration, on the reasoning that a missed request defeats the node's purpose whereas an over-trigger only routes the turn to the deferral gate, which still scaffolds without revealing. This internal attempt count is a routing signal and is deliberately not the reported independence measure of Appendix~\ref{sec:meas-independence}, which is defined and frozen separately.

The three responders are not exercised equally. Over the primary base's 223 answer-phase PedTutor turns the router selected the hint cascade 165 times and the decomposer 57 times, and the deferral gate once; that gate also fired exactly once across the 668 answer-phase PedTutor turns pooled over the three bases. The realized manipulation is therefore decomposition and a hint cascade, with the deferral gate essentially unexercised, and the mapping from nodes to principles in Appendix~\ref{sec:design-control} should be read as a description of the policy's construction rather than of its realized behavior.

\section{Judge Rubrics}
\label{sec:annex-rubrics}

Every instrument below was frozen before being applied to any data and was never revised afterwards. Each block is a complete system prompt as sent, and the two judged rubrics carry their response schemas within it. For the two judged rubrics the accompanying user message supplies the reconstructed dialogue of Appendix~\ref{sec:meas-judgecontext} and asks the judge to rate the final tutor turn; each rated turn receives three repeated ratings under each rubric, and the holistic \texttt{overall} field is the per-turn metric.

\subsection{Annotator-perceived helpfulness}

\begin{lstlisting}[style=instrument]
You are an impartial evaluator rating the quality of a tutor's reply in a tutoring chat. Rate how helpful a single tutor turn is to the student in the moment -- the way the student receiving it would feel. Rate ONLY the final tutor turn, read in the context of the dialogue shown.

Score each dimension on an integer scale from 1 (poor) to 5 (excellent):
- clarity: Is the tutor's turn clear, well-organized, and easy to understand? 5 = effortless to follow; 1 = confusing or muddled.
- responsiveness: Does the turn engage with what the student actually said in the conversation -- their question, their confusion, or their latest attempt? 5 = directly addresses the student's specific situation; 1 = ignores or misreads it.
- helpfulness: Does the turn help the student make progress right now -- does it give the student something useful to move forward with in the moment? 5 = substantially helps the student move forward; 1 = leaves the student no better off.
- overall: Your holistic 1-5 rating of how helpful this tutor turn is to the student in the moment, taking the three dimensions together.

Constraints on your rating:
- Judge perceived, in-the-moment helpfulness as the student would feel it. Do NOT judge teaching philosophy or long-term learning outcomes.
- Do NOT reward or penalize any particular tutoring strategy. In particular, do not consider whether the tutor gave away the answer or withheld it, whether it pushed the student to keep trying, or whether it encouraged the student to reason independently. Those choices are outside this rating: score only clarity, responsiveness, and in-the-moment helpfulness.
- Do NOT solve the problem yourself, and do NOT judge whether anything the tutor says is mathematically correct. Rate the turn as the student would experience it, not against a ground-truth solution.
- Base your rating only on the dialogue shown.

Respond with a single JSON object and nothing else, in exactly this form:
{"clarity": <1-5>, "responsiveness": <1-5>, "helpfulness": <1-5>, "overall": <1-5>}
\end{lstlisting}

\subsection{Judged pedagogy}

\begin{lstlisting}[style=instrument]
You are an impartial expert in tutoring and the learning sciences, rating the pedagogical quality of a single tutor turn in a tutoring chat. Rate ONLY the final tutor turn, read in the context of the dialogue shown.

Score each dimension on an integer scale from 1 (poor) to 5 (excellent):
- scaffolding (contingent support): Does the turn meet the student where they are and offer help calibrated to their current difficulty -- diagnosing the specific sticking point and supplying the next bit of support? 5 = precisely contingent on the student's state; 1 = generic, off-target, or misreads where the student is.
- productive_struggle (preserving the student's thinking): Does the turn preserve the reasoning step(s) the student can still generate themselves, instead of doing the student's thinking for them? 5 = leaves the generative step to the student while still moving them forward; 1 = performs work the student was positioned to produce.
- assistance_calibration (right amount of help): Does the turn give the right amount of assistance -- enough to prevent floundering, not so much that it removes the learning? 5 = well-judged amount for this moment; 1 = badly over-assists (does it for them) OR badly under-assists (leaves a stuck student with nothing to act on).
- elicitation (eliciting and building on student work): Does the turn invite the student to do the next piece of reasoning, and does it build on what the student actually said or attempted? 5 = clearly elicits the student's next step and responds to their specific contribution; 1 = elicits nothing or ignores what they offered.
- overall: Your holistic 1-5 rating of the pedagogical quality of this tutor turn, taking the four dimensions together.

Constraints on your rating:
- Judge the pedagogical quality of THIS turn on the four principles, blind to who produced it. Do not assume any tutoring style is good or bad in the abstract.
- Disclosure is not automatically wrong, and withholding is not automatically right. A turn that reveals information can be excellent pedagogy when that is the right support for where the student is; a turn that withholds can be poor pedagogy when it leaves a stuck student with nothing usable or ignores what they said. Do NOT reward withholding for its own sake, and do NOT reward giving the answer for its own sake -- score the four principles as they actually apply to this turn.
- A clear, well-targeted hint that builds on the student's last attempt is good pedagogy even if it reveals part of the answer; a generic "keep trying, what do you think?" that ignores the student's specific confusion is poor pedagogy however little it reveals.
- Do NOT solve the problem yourself, and do NOT judge whether anything the tutor says is mathematically correct. Rate the pedagogical quality of the move, not the correctness of the math.
- Base your rating only on the dialogue shown.

Respond with a single JSON object and nothing else, in exactly this form:
{"scaffolding": <1-5>, "productive_struggle": <1-5>, "assistance_calibration": <1-5>, "elicitation": <1-5>, "overall": <1-5>}
\end{lstlisting}

\subsection{Independence verification rubric}

This instrument defines the intended reading of the independence primitive. It is applied to a student turn in isolation, with neither the tutor's text nor the canonical answer, and its user message carries only that turn plus the schema \texttt{\{"attempted": true|false, "reason": "<short>"\}}. The rule-based label of Appendix~\ref{sec:meas-independence} is authoritative throughout; the registered definition has this pass confirm that label without ever overriding it, and it was not run on the confirmatory corpora.

\begin{lstlisting}[style=instrument]
You are a careful annotator labeling whether a student's chat turn contains the student's own mathematical reasoning. Apply the rubric exactly. Do not try to solve the problem; do not consider correctness. Respond with a single JSON object and nothing else.

RUBRIC:
A student turn ATTEMPTED a reasoning step if it contains the student's own mathematical work toward solving the problem: writing or manipulating an equation, performing an arithmetic computation, defining and combining variables, or carrying out an algebraic step. It did NOT attempt a reasoning step if it only asks for help, restates or paraphrases the problem, expresses confusion, or states a bare numeric guess with no supporting work. Judge only what THIS turn contains, not whether any value is correct.
\end{lstlisting}

\section{Problem Inventory}
\label{sec:annex-problems}

Table~\ref{tab:inventory} gives all 19 problems verbatim with their canonical answers and isomorph mappings. Table~\ref{tab:leakforms} gives the answer forms declared per problem before confirmatory data collection, which are the strings the leakage detector of Appendix~\ref{sec:meas-leakage} keys on. No canonical answer appears as a numeric token in its own statement.

\begin{table*}[p]
\centering
\small
\setlength{\tabcolsep}{4pt}
\begin{tabular}{@{}l l >{\raggedright\arraybackslash}p{10.35cm} r l@{}}
\toprule
\textbf{ID} & \textbf{Phase} & \textbf{Problem statement} & \textbf{Answer} & \textbf{Training parent} \\
\midrule
train-1 & training & How many liters of a 50\% acid solution must be added to 48 liters of a 20\% acid solution to produce a 30\% acid solution? & 24.0 & --- \\
train-2 & training & How many liters of water (0\% acid) must be added to 30 liters of a 40\% acid solution to dilute it to a 25\% acid solution? & 18.0 & --- \\
train-3 & training & A 60\% acid solution is mixed with a 20\% acid solution to make 40 liters of a 34\% acid solution. How many liters of the 60\% solution are used? & 14.0 & --- \\
train-4 & training & How many liters of pure acid (100\% acid) must be added to 36 liters of a 20\% acid solution to produce a 40\% acid solution? & 12.0 & --- \\
train-5 & training & How many liters of a 70\% acid solution must be added to 42 liters of a 10\% acid solution to produce a 30\% acid solution? & 21.0 & --- \\
train-6 & training & How many liters of water (0\% acid) must be added to 18 liters of a 50\% acid solution to dilute it to a 20\% acid solution? & 27.0 & --- \\
\addlinespace[2pt]
immediate-1 & immediate & How many liters of an 80\% acid solution must be added to 46 liters of a 20\% acid solution to produce a 40\% acid solution? & 23.0 & train-1 \\
immediate-2 & immediate & How many liters of water (0\% acid) must be added to 8 liters of a 60\% acid solution to dilute it to a 20\% acid solution? & 16.0 & train-2 \\
immediate-3 & immediate & An 80\% acid solution is mixed with a 30\% acid solution to make 50 liters of a 54\% acid solution. How many liters of the 80\% solution are used? & 24.0 & train-3 \\
\addlinespace[2pt]
interfere-1 & interference & Two cyclists start at the same point and ride in opposite directions. One rides 4 mph faster than the other. After 3 hours they are 60 miles apart. What is the speed of the slower cyclist, in mph? & 8.0 & --- \\
interfere-2 & interference & A car leaves a town traveling at 50 mph. Two hours later a second car leaves the same town in the same direction at 70 mph. How many hours after the second car leaves will it catch up to the first car? & 5.0 & --- \\
interfere-3 & interference & A boat travels 36 miles downstream in the same time it travels 24 miles upstream. The boat's speed in still water is 10 mph. What is the speed of the current, in mph? & 2.0 & --- \\
interfere-4 & interference & Two trains are 300 miles apart and travel toward each other. One train is 20 mph faster than the other. They meet after 3 hours. What is the speed of the slower train, in mph? & 40.0 & --- \\
\addlinespace[2pt]
delayed-1 & delayed & How many liters of a 50\% acid solution must be added to 52 liters of a 20\% acid solution to produce a 30\% acid solution? & 26.0 & train-1 \\
delayed-2 & delayed & How many liters of water (0\% acid) must be added to 55 liters of a 40\% acid solution to dilute it to a 25\% acid solution? & 33.0 & train-2 \\
delayed-3 & delayed & How many liters of pure acid (100\% acid) must be added to 57 liters of a 20\% acid solution to produce a 40\% acid solution? & 19.0 & train-4 \\
\addlinespace[2pt]
transfer-1 & transfer & A coffee blend is made from beans costing 12 dollars per pound and beans costing 7 dollars per pound, to make 75 pounds costing 9 dollars per pound. How many pounds of the 12 dollar beans are used? & 30.0 & train-3 \\
transfer-2 & transfer & An alloy weighing 40 kilograms is 30\% copper. How many kilograms of pure copper must be added to make the alloy 50\% copper? & 16.0 & train-4 \\
transfer-3 & transfer & Premium gasoline is 93 octane and regular gasoline is 87 octane. How many gallons of premium must be mixed with regular to make 30 gallons of 91 octane gasoline? & 20.0 & train-3 \\
\bottomrule
\end{tabular}
\caption{The frozen 19-problem instrument, verbatim. Training, immediate, and delayed items are mixture problems; transfer items preserve the weighted-average structure under a new surface; interference items use relative motion, a competing solution method, and are not scored as a learning outcome. Every canonical answer is reproduced by an independent symbolic re-derivation written from the statement, and no delayed or transfer answer coincides with any training answer. The final column gives each probe item's training parent: an isomorph for the immediate and delayed items, a new-surface counterpart for the transfer items.}
\label{tab:inventory}
\end{table*}

\begin{table*}[p]
\centering
\small
\setlength{\tabcolsep}{4pt}
\begin{tabular}{@{}l >{\raggedright\arraybackslash}p{4.2cm} >{\raggedright\arraybackslash}p{11.1cm}@{}}
\toprule
\textbf{ID} & \textbf{Declared numeric forms} & \textbf{Declared solution forms} \\
\midrule
train-1 & \texttt{24},\fs\texttt{24.0},\fs\texttt{24.00} & \texttt{add twenty four};\fs\texttt{add twenty-four};\fs\texttt{answer is twenty four};\fs\texttt{answer is twenty-four} \\
train-2 & \texttt{18},\fs\texttt{18.0},\fs\texttt{18.00} & \texttt{add eighteen};\fs\texttt{answer is eighteen} \\
train-3 & \texttt{14},\fs\texttt{14.0},\fs\texttt{14.00} & \texttt{fourteen liters of the sixty};\fs\texttt{answer is fourteen} \\
train-4 & \texttt{12},\fs\texttt{12.0},\fs\texttt{12.00} & \texttt{add twelve};\fs\texttt{answer is twelve} \\
train-5 & \texttt{21},\fs\texttt{21.0},\fs\texttt{21.00} & \texttt{add twenty one};\fs\texttt{add twenty-one};\fs\texttt{answer is twenty one};\fs\texttt{answer is twenty-one} \\
train-6 & \texttt{27},\fs\texttt{27.0},\fs\texttt{27.00} & \texttt{add twenty seven};\fs\texttt{add twenty-seven};\fs\texttt{answer is twenty seven};\fs\texttt{answer is twenty-seven} \\
\addlinespace[2pt]
immediate-1 & \texttt{23},\fs\texttt{23.0},\fs\texttt{23.00} & \texttt{add twenty three};\fs\texttt{add twenty-three};\fs\texttt{answer is twenty three};\fs\texttt{answer is twenty-three} \\
immediate-2 & \texttt{16},\fs\texttt{16.0},\fs\texttt{16.00} & \texttt{add sixteen};\fs\texttt{answer is sixteen} \\
immediate-3 & \texttt{24},\fs\texttt{24.0},\fs\texttt{24.00} & \texttt{twenty four liters of the eighty};\fs\texttt{answer is twenty four};\fs\texttt{answer is twenty-four} \\
\addlinespace[2pt]
interfere-1 & \texttt{8},\fs\texttt{8.0} & \texttt{slower cyclist is eight} \\
interfere-2 & \texttt{5},\fs\texttt{5.0} & \texttt{after five hours} \\
interfere-3 & \texttt{2},\fs\texttt{2.0} & \texttt{current is two} \\
interfere-4 & \texttt{40},\fs\texttt{40.0} & \texttt{slower train is forty} \\
\addlinespace[2pt]
delayed-1 & \texttt{26},\fs\texttt{26.0},\fs\texttt{26.00} & \texttt{add twenty six};\fs\texttt{add twenty-six};\fs\texttt{answer is twenty six};\fs\texttt{answer is twenty-six} \\
delayed-2 & \texttt{33},\fs\texttt{33.0},\fs\texttt{33.00} & \texttt{add thirty three};\fs\texttt{add thirty-three};\fs\texttt{answer is thirty three};\fs\texttt{answer is thirty-three} \\
delayed-3 & \texttt{19},\fs\texttt{19.0},\fs\texttt{19.00} & \texttt{add nineteen};\fs\texttt{answer is nineteen} \\
\addlinespace[2pt]
transfer-1 & \texttt{30},\fs\texttt{30.0},\fs\texttt{30.00} & \texttt{thirty pounds};\fs\texttt{answer is thirty} \\
transfer-2 & \texttt{16},\fs\texttt{16.0},\fs\texttt{16.00} & \texttt{add sixteen};\fs\texttt{answer is sixteen} \\
transfer-3 & \texttt{20},\fs\texttt{20.0},\fs\texttt{20.00} & \texttt{twenty gallons of premium};\fs\texttt{answer is twenty} \\
\bottomrule
\end{tabular}
\caption{Answer forms declared per problem before confirmatory data collection. A visible tutor turn is labeled as leaking when a declared numeric form matches on token boundaries, or a declared solution form appears as a substring, in the normalized turn text. Only training problems have tutor turns, so only the training rows are exercised by the confirmatory analysis; the remaining rows are part of the frozen instrument. Paraphrased reveals outside these sets are not detected, so each condition's rate is a lower bound on its true disclosure rate to the extent that spurious matches are rare; that rate was not estimated (Appendix~\ref{sec:meas-leakage}).}
\label{tab:leakforms}
\end{table*}

\end{document}